\documentclass[11pt]{article}
\usepackage{geometry}                
\usepackage{setspace}
\usepackage[parfill]{parskip}    
\usepackage{newtxtext}

\usepackage{amsmath}
\usepackage{amsthm}
\usepackage{amssymb}
\usepackage{mathtools}
\usepackage{bm}

\newtheorem{theorem}{Theorem}

\newtheorem{definition}{Definition}

\usepackage{graphicx}
\usepackage{subfig}
\usepackage{booktabs}       
\newcommand{\tabincell}[2]{\begin{tabular}{@{}#1@{}}#2\end{tabular}} 

\usepackage[hidelinks]{hyperref}
\usepackage{url}

\usepackage{color}

\title{Recent advances in deep learning theory}

\author{Fengxiang He\thanks{The authors were with UBTECH Sydney AI Centre, School of Computer Science, Faculty of Engineering, the University of Sydney, Darlington NSW 2008, Australia. Email: \href{mailto:fengxiang.f.he@gmail.com}{fengxiang.f.he@gmail.com} and \href{mailto:dacheng.tao@sydney.edu.au}{dacheng.tao@sydney.edu.au}.} \and Dacheng Tao\footnotemark[1]}

\date{}

\begin{document}

\maketitle

\begin{abstract}
Deep learning is usually described as an experiment-driven field under continuous criticizes of lacking theoretical foundations. This problem has been partially fixed by a large volume of literature which has so far not been well organized. This paper reviews and organizes the recent advances in deep learning theory. The literature is categorized in six groups: (1) complexity and capacity-based approaches for analyzing the generalizability of deep learning; (2) stochastic differential equations and their dynamic systems for modelling stochastic gradient descent and its variants, which characterize the optimization and generalization of deep learning, partially inspired by Bayesian inference; (3) the geometrical structures of the loss landscape that drives the trajectories of the dynamic systems; (4) the roles of over-parameterization of deep neural networks from both positive and negative perspectives; (5) theoretical foundations of several special structures in network architectures; and (6) the increasingly intensive concerns in ethics and security and their relationships with generalizability.
\end{abstract}

~~~~~~~~~~\textbf{Keywords:} neural networks, deep learning, deep learning theory.

\section{Introduction}
\label{sec:introduction}

{\it Deep learning} can be broadly defined as a family of algorithms employing {\it artificial neural networks} to discover knowledge from {\it experience} for predictions or decision-making \cite{lecun2015deep}. Canonical forms of the experience can be human-annotated electronic records as a dataset or the interactions between the learner or the electronic environment depending on the scenarios \cite{mohri2018foundations}. A normal artificial neural network in deep learning usually connects a sequence od {\it weight matrix} followed by a {\it nonlinear activation function} as a network, which is typical of a considerably large parameter size.

The term of {\it deep learning} was introduced to machine learning by Dechter \cite{dechter1986learning} and then to brain-inspired algorithms by Aizenberg {\it et al.} \cite{aizenberg2000multi}, while several major concepts wherein can be dated back to early 1940s. The research of deep learning saw two rises and then falls in 1940s-1960s \cite{mcculloch1943logical, hebb1949organization, rosenblatt1958perceptron} and 1980s–1990s \cite{rumelhart1986learning}. The third and current wave starts in 2006 \cite{bengio2006greedy, hinton2006fast, ranzato2006efficient} and lasts till now. The recent wave has substantially reshaped many real-world application domains, including computer vision \cite{he2016deep}, natural language processing \cite{deng2018deep, otter2020survey}, speech processing \cite{deng2014deep}, 3D point cloud processing \cite{guo2020deep}, data mining \cite{witten2016data}, recommender system \cite{zhang2019deep}, autonomous vehicles \cite{litjens2017survey, sun2006road}, medical diagnosis \cite{kulikowski1980artificial, silver2016mastering}, and drug discovery \cite{chen2018rise}.

To date, however, the development of deep learning heavily relies on experiments but without solid theoretical foundations. Many facets of the mechanism of deep learning still remain unknown. We are continuously surprised by finding heuristic approaches can achieve excellent performance across extensive areas, though sometimes also considerably instable. Meanwhile, the suggested intuitions are usually left un-verified or even un-checked. Such practice is tolerated and has become common in deep learning research. This black-box nature introduces unknown risks to deep learning applications. This unawareness considerably undermines our capabilities in identifying, managing, and preventing the algorithm-led disasters, and further severely hurts the confidence in applying the recent progress to many industrial sectors, especially those in security-critical areas, such as autonomous vehicles, medical diagnosis, and drug discovery. This also backlashes on the future development of innovative deep learning algorithms designing.

A major part of potential theoretical foundations is the {\it generalizability}, which refers to the capability of models well trained on the training data by deep learning algorithms predicting on the unseen data \cite{vapnik2013nature, mohri2018foundations}. Since the training data can not cover every future circumstance, good generalizability secures the learned model to be able to handle unseen events. This is particularly important where long-tail events constantly appear and are possible to cause fatal disasters.

Statistical learning theory has established theories for the generalizability relying on the hypothesis complexity \cite{vapnik2013nature, mohri2018foundations}. Could these tools solve the problem in deep learning theory? The answer is no in general. Conventional tools usually develops generalization bounds based on hypothesis complexities, such as VC-dimension \cite{blumer1989learnability, vapnik2006estimation}, Rademacher complexity \cite{koltchinskii2000rademacher, koltchinskii2001rademacher, bartlett2002rademacher}, and covering number \cite{dudley1967sizes, haussler1995sphere}. These complexities, in classic results, heavily rely on the model sizes. This introduces the {\it Occam's razor} principle:
\begin{quotation}
\centering
{\it Plurality should not be posited without necessity;}
\end{quotation}
{\it i.e.}, one needs to find a model sufficiently small to prevent overfitting as long as the model can fit the training sample. However, deep learning models usually have extremely large model sizes, which sometimes makes the generalization bound even larger than the potential maximal value of the loss function. Moreover, the Occam's razor principle suggests a positive correlation between the generalizability and the model size, which is no longer observed in deep learning. In contrast, deeper and wider networks often have advantaged performance \cite{canziani2016analysis}. The irreconciliation between the excellent generalizability of deep learning and its extreme over-parameterization is like a ``cloud'' to conventional complexity-based learning theory.

\subsection{A brief overview of the status quo}

Early works have attempted to establish the theoretical foundations of deep learning \cite{musavi1994generalization, goldberg1995bounding, bartlett1996vc, bartlett1999almost, baum1989size, maass1997bounds, anthony2009neural} but largely stagnated with the dip of the wider deep learning research.

The recent story started from the work by Zhang {\it {\it et al.}} \cite{zhang2017understanding} in 2017. The authors conducted systematic experiments to explore the generalization ability of deep neural networks. They show that neural networks can almost perfectly fit the training data even when the training labels are random. This paper attracts the community of learning theory to the important topic that how to theoretically interpret the success of deep neural networks. Kawaguchi {\it et al.} \cite{kawaguchi2017generalization} discuss many open problems regarding the excellent generalization ability of deep neural networks despite the large capacity, complexity, possible algorithmic instability, non-robustness, and sharp minima. The authors also provide some insights to solve the problems. Since then, the importance of deep learning theory has been widely recognized. A large volume of literature has emerged to establish theoretical foundations of deep learning. In this paper, we review the related literature and organize them into following six categories.

\textbf{Complexity and capacity-based method for analyzing the generalizability of deep learning.} Conventional statistical learning theory has established the a series of upper bounds on the generalization error (generalization bounds) based on the complexity of hypothesis space, such as VC-dimension \cite{blumer1989learnability, vapnik2006estimation}, Rademacher complexity \cite{koltchinskii2000rademacher, koltchinskii2001rademacher, bartlett2002rademacher}, and covering number \cite{dudley1967sizes, haussler1995sphere}. Usually, these generalization bounds explicitly rely on the model size. They suggest that controlling the model size can help models generalize better. However, the colossal model sizes of deep learning models also make the generalization bounds vacuous. It is thus much desired if we can develop size-independent hypothesis complexity measure and generalization bounds. A promising idea is to characterize the complexity of the ``effective'' hypothesis space which can be learned in deep learning. The effective hypothesis space can be significantly smaller than the whole hypothesis space. We thereby can expect to obtain a much smaller generalization guarantee.

\textbf{Stochastic differential equations (SDE) for modelling stochastic gradient descent (SGD) and its variants, which dominate the optimization algorithms in deep learning.} The dynamic systems of these SDEs determine the trajectory of the weights in training neural networks and their steady distributions represent the learned networks. Through the SDEs and their dynamics, many works establish guarantees for the optimization and generalization of deep learning. The ``effective'' hypothesis space is exactly the hypothesis space ``that can be found by SGD''. Thus, studying the generalizability of deep learning via SGD would be straightforward. Moreover, this family of methods are partially inspired by Bayesian inference. This echos the previous story of variation inference which solves Bayesian inference in an optimization manner in order to address the scaling issue. This interplay between stochastic gradient methods and Bayesian inference would help advance both areas.

\textbf{The geometrical structures of the highly complex empirical risk landscape that drive the trajectories of the dynamic systems.} The geometry of the loss landscape plays major roles in driving the trajectories of the SDEs: (1) the derivatives of the loss are components of the SDEs; and (2) the loss would serve as boundaries conditions of the SDEs. Therefore, understanding the loss surface would be a key step in establishing theoretical foundations for deep learning.

In convention, the learnability and optimization ability are usually guaranteed in ``regularized'' problems.\footnote{It is worth noting that the ``regularized'' here or the ``regularization'' later refer to the approaches to make ill-posed problems solvable. It is not necessarily relating to adding ``regularizers''.} The ``regularization'' can be characterized by a variety of terms, including convexity, Lipschitz continuity, and differentiability. However, these factors are no longer secured in deep learning, at least not obviously. Neural networks usually consist of a large number of nonlinear activations. The nonlinearity in activations makes the loss surface extremely non-smooth and non-convex. The established guarantees for convex optimization become invalid. The prohibitive complexity of loss surface checks the community touching the geometry of the loss landscape and even deep learning theory for a long time. However, the complex geometry of the loss surface exactly characterizes the behavior of deep learning. It is the ``high way'' to understand deep learning via its loss surface.

\textbf{The roles of over-parameterization of deep neural networks.} Over-parameterization is usually treated as a major barrier in developing meaningful generalization bounds for deep learning via complexity-based approaches. However, recent studies suggest that over-parameterization would make major contributes in shaping the loss surface of deep learning - making the loss surface more smooth and even ``analogous" convex. Moreover, many works have proven that neural networks in the extreme over-parameterization regime are equivalent to some simpler models, such as Gaussian kernels.

\textbf{Theoretical foundations for several special structures in network architectures.} In the previous review, we focus on the result generally stand for all neural networks. Meanwhile, deep neural networks designing involves many special techniques. These structures also contribute significantly to deep learning's excellent performance. We review the theoretical results on convolutional neural networks, recurrent neural networks, and networks for permutation invariant/equivariant functions.

\textbf{The intensive concerns in ethics and security and their relationships with deep learning theory.} Deep learning has been deployed in an increasingly wide spectrum of application domains. Some of them involve highly private personal data, such as images and videos on mobile phones, health data, and final records. Some other scenarios may require deep learning to deliver highly sensitive decision-making, such as mortgage approval, college admission, and credit assessment. Moreover, deep learning models have been shown vulnerable to adversarial examples. How to secure a deep learning system from breaches related to privacy preservation, fairness concerning, and adversarial attacks are of significant importance.

\subsection{Overview of this paper}

This paper reviews the recent advances in establishing the theoretical foundations of deep learning. We acknowledge that a few papers have reviewed deep learning theory. Alom {\it et al.} \cite{alom2019state} give a survey of the techniques used in deep learning. Sun \cite{sun2019optimization} reviews the theory of optimization in deep learning. E {\it et al.} \cite{e2020towards} summarize results and challenges related to approximation and Rademacher complexity, loss surface, and the convergence and implicit regularization of optimization in deep learning. Our survey is the most comprehensive. We organize the literature in a unique view and provide new insights for future works.

The excellent generalizability of deep learning is like a ``cloud'' to conventional complexity-based learning theory: the over-parameterization of deep learning makes almost all existing tools vacuous. Existing works try to tackle this problem through three major paths: (1) developing size-independent complexity measures, which can characterize the complexity of the “effective” hypothesis space that can be learned, instead of the whole hypothesis space. The related works are discussed in Section \ref{sec:complexity}; and (2) modelling the learned hypothesis through stochastic gradient methods, the dominant optimizers in deep learning, based on the stochastic differential functions and the geometry of the associated loss functions. The related works are reviewed in Sections \ref{sec:SGM} and \ref{sec:loss}; and (3) over-parameterization surprisingly bring many good properties to the loss functions and further secures the optimization and generalization performances. The related works are given in Section \ref{sec:over-parameterization}. Parallel to these, Section \ref{sec:special_structures} reviews the theoretical foundations on special structures of network architecture.

Another important facet of machine learning is the rising concern on the ethical and security issues, including privacy preservation, adversarial robustness, and fairness protection. Specifically, privacy preservation and adversarial robustness have been discovered closely related to generalizability: a good generalizability usually means a good privacy-preserving ability; and more robust algorithms might have. Efforts of understanding the interplay between these issues in the scenario of deep learning are also discussed in this paper. The related works are discussed in Section \ref{sec:ethics_security}.

\subsection{Notations}

Suppose $S = \{(x_1, y_1), \ldots, (x_N, y_N) | X_i \in \mathcal X \subset \mathbb R^{d_X}, Y_i \in \mathcal Y \subset \mathbb R^{d_Y}, i = 1, \ldots, N\}$ is a training sample set, where $d_X$ and $d_Y$ are the dimensions of the feature $X$ and the label $Y$, respectively. For the brevity, we define $z_i = (x_i, y_i) \in \mathcal Z = \mathcal X \times \mathcal Y$. All $z_i$ are independent and identically distributed (i.i.d.) observations of the random variable $Z = (X, Y) \in \mathcal Z$. 
Machine learning algorithms are designed to learn hypothesis $h$ from the training data; all potential hypotheses constitute a hypothesis space $\mathcal{H}$. 

A neural network is usually defined as the following form,
\begin{equation*}
x \mapsto W_D \sigma_{D-1} (W_{D-1} \sigma_{D-2} (\ldots \sigma_1(W_1 x))),
\end{equation*}
where $D$ is the depth of the neural network, $W_j$ is the $j$-th weight matrix, and $\sigma_j$ is the $j$-th nonlinear activation function. Usually, we assume the activation function $\sigma_j$ is a continuous function between Euclidean spaces and $\sigma_j(0) = 0$. Popular activations in practice including softmax, sigmod, and tanh functions. 

Generalization bound measures the generalization ability of an algorithm. For any hypothesis $h$ learned by an algorithm $\mathcal A$, the expected risk $\mathcal R(h)$ and empirical risk $\hat{\mathcal R}_S(h)$ with respect to the training sample set $S$ are respectively defined as follows,
\begin{gather*}
	\mathcal R(h) = \mathbb E_Z l(h, Z),~
	\hat{\mathcal R}_S(h) = \frac{1}{N} \sum_{i=1}^N l(h, z_i).
\end{gather*}
Additionally, when the output hypothesis $h$ of algorithm $\mathcal{A}$ is stochastic, we usually calculate the expectations of the expected risk $\mathcal R(h)$ and empirical risk $\hat{\mathcal R}(h)$ with respect to the randomness introduced by the algorithm $\mathcal A$ as follows,
\begin{gather*}
	\mathcal R(\mathcal{A}) = \mathbb E_{\mathcal A(S)} \mathcal R(\mathcal A(S)),~
	\hat{\mathcal R}_S(\mathcal{A}) = \mathbb{E}_{\mathcal A(S)} \hat{\mathcal R}_S(\mathcal A(S)),
\end{gather*}
where $\mathcal A(S)$ is the hypothesis learned by the algorithm $\mathcal A$ on the sample $S$. Then, the generalization error can be defined as the difference between the expected risk and empirical risk.

\section{Capacity and complexity: Advances and predicament}
\label{sec:complexity}

Conventional statistical learning theory establishes generalization guarantees based on the complexity of the hypothesis space. Following this line, some works have proposed upper bounds of the generalization error of deep learning.

\subsection{Worst-case bounds based on VC-dimension}

A major measure for evaluating the hypothesis complexity in conventional statistical learning theory is the Vapnik-Chervonenkis dimension (VC-dimension, \cite{blumer1989learnability, vapnik2006estimation}) which is defined as follows.

\begin{definition}[growth function, shattering, and VC-dimension]
For any non-negative integer $m$, the growth function of hypothesis space $\mathcal H$ is defined as follows,
\begin{equation*}
\Pi_{\mathcal H} (m) \coloneqq \max_{x_1, \ldots, x_m \in \mathcal X} | \{ (h(x_1), \ldots, h(x_m)): h \in \mathcal H \} |.
\end{equation*}
If $\Pi_{\mathcal H} (m) = 2^m$, we say $\mathcal H$ shatters the dataset $\{x_1, \ldots, x_m \in \mathcal X\}$. We define the VC-dimension $\text{VCdim}(\mathcal H)$ as the largest size of shattered set.
\end{definition}

One can obtain a uniform generalization bound via the VC-dimension as follows. 

\begin{theorem}
Assume hypothesis space $\mathcal H$ has VC-dimension $\text{VCdim}(\mathcal H)$. Then, for any $\delta > 0$, with probability $1 - \delta$, the following inequality holds for any $h \in \mathcal H$,
\begin{equation*}
\mathcal R(h) \le \hat{\mathcal R} (h) + \sqrt{\frac{2 \text{VCdim}(\mathcal H) \log \frac{em}{\text{VCdim}(\mathcal H)}}{m}} + \sqrt{\frac{\log \frac{1}{\delta}}{2m}},
\end{equation*}
where $m$ is the training sample size.
\end{theorem}

Thus, results on the VC-dimension of neural networks can characterize its generalizability. Goldberg and Jerrum \cite{goldberg1995bounding} gives an $\mathcal O(W^2)$ upper bound for the VC dimension of neural networks with parameter size $W$ and depth $D$. It was improved to $\mathcal O(W \log (WD))$ by Bartlett and Williamson \cite{bartlett1996vc} and then to $\mathcal O(W L \log W +W L^2)$ by Bartlett {\it et al.}\cite{bartlett1999almost}. Other on VC-dimension were also given by Baum and Haussler \cite{baum1989size} and Maass \cite{maass1997bounds}.

The tightest upper bound for the VC-dimension so far is proven by Harvey {\it et al.} \cite{harvey2017nearly} as follows,

\begin{theorem}
 Consider a neural network with $W$ parameters and $U$ units with activation functions that are piecewise polynomials with at most $p$ pieces and of degree at most $d$. Let $F$ be the set of (real-valued) functions computed by this network. Then,
\begin{equation*}
\text{VCdim}(\text{sgn}(\mathcal F)) = \mathcal O(W U \log((d + 1)p)).
\end{equation*}
\end{theorem}

However, such upper bounds on the VC-dimensions all heavily rely on the model size $W$, which is extremely large in deep learning. In some cases, the generalization bound can be significantly larger than the largest potential value of the loss function (such as $1$ for the $0$-$1$ loss).

Harvey {\it et al.} \cite{harvey2017nearly} also give a lower bound for the VC-dimension as the following theorem.


\subsection{Margin bounds: From exponential depth dependence to depth independence}

Margin bounds are another family of generalization bounds \cite{vapnik2006estimation, schapire1998boosting, bartlett1999generalization, koltchinskii2002empirical, taskar2004max}. Compared with the worst-case bounds based on the VC-dimension, margin bounds deliver strong guarantees for the learned models that they can achieve small empirical margin loss for large confidence margin. Similar generalization guarantees can be obtained via the luckiness \cite{schapire1998boosting, koltchinskii2002empirical}.

\textbf{Margin bound.}  For any distribution $\mathcal D$ and margin $\gamma > 0$, we define the expected margin loss for hypothesis $h$ as follows.
\begin{equation*}
L_\gamma (h) = \mathbb P_{(x, y) \sim \mathcal D} \left( h(x)[y] \le \gamma + \max_{j \neq y} h(x)[j] \right),
\end{equation*}
where $h(x)[j]$ is the $j$-th component of the vector $h(x)$. The generalization bounds under the margin loss are called margin bounds.

One can prove a margin bound via the the {\it covering number}  \cite{dudley1967sizes, haussler1995sphere} or {\it Rademacher complexity} \cite{koltchinskii2000rademacher, koltchinskii2001rademacher, bartlett2002rademacher} of the hypothesis space defined as follows.

\begin{definition}[covering number]
The covering number of covering number $\mathcal N(\mathcal H, \epsilon, \| \cdot \|)$ of space $\mathcal H$ is defined to be the least cardinality of any subset $V \subset \mathcal H$ that covers $\mathcal H$ at scale $\epsilon$ under metric $\| \cdot \|$; {\it i.e.}, 
$\sup_{A \in \mathcal H} \min_{B \in V} \| A - B \| \le \epsilon$. 
\end{definition}


\begin{definition}[empirical Rademacher complexity and Rademacher complexity]
Given a real-valued function class $\mathcal H$ and a dataset $S$, the empirical Rademacher complexity is defined as follows,
\begin{equation*}
\hat{\mathfrak R}(\mathcal H) = \mathbb E_\epsilon \left[ \sup_{h \in \mathcal H} \frac{1}{m} \sum_{i = 1}^m \epsilon_i h(x_i) \right],
\end{equation*}
where $\epsilon = \{\epsilon_1, \ldots, \epsilon_m\}$ is a random vector distributed in the space $\{-1, +1\}^m$. Further, we can define Rademacher complexity as follows,
\begin{equation*}
{\mathfrak R}(\mathcal H) = \mathbb E_{S} \hat{\mathfrak R}(\mathcal H).
\end{equation*}
\end{definition}

Intuitively, covering number indicates how many balls are needed to cover the hypothesis space; and Rademacher complexity measures the maximal correlations between the output of a hypothesis and a noise vector, and thus characterizes the ``goodness-of-fit" of the hypothesis space to a noise. Apparently, a larger covering number or Rademacher complexity corresponds to a larger hypothesis complexity. 

One can also upper bound the Rademacher complexity via the covering number through the Dudley entropy integral \cite{mohri2018foundations} as follows.

\begin{theorem}[Dudley entropy integral]
Let $\mathcal H$ be a real-valued function class taking values in $[0, 1]$, and assume that $0 \in \mathcal H$. Then,
\begin{align*}
& \hat{\mathfrak R}(\mathcal H) 
 \le \inf_{\alpha > 0} \left( \frac{4\alpha}{\sqrt{m}} + \frac{12}{m} \int_{\alpha}^{\sqrt{m}} \sqrt{\log \mathcal N(\mathcal H, \epsilon, \| \cdot \|_2)} \text{d}\epsilon \right).
\end{align*}
\end{theorem}

Besides, one can obtain margin bounds via PAC-Bayesian theory as below, which corporates PAC theory and Bayesian statistics \cite{mcallester1999pac, mcallester1999some}.

\begin{theorem}[PAC-Bayesian theory]
\label{thm:PAC-Bayesian}
Suppose the prior and the posterior of the model parameter $\theta$ is $\mathcal P$ and $\mathcal Q$. Then,
\begin{equation}
\label{eq:PAC-Bayesian}
\mathbb E_\theta \mathcal R(h(\theta)) = \mathbb E_\theta \hat{\mathcal R}(h(\theta)) + \sqrt{\frac{\text{KL}(\mathcal Q || \mathcal P) + \log \frac{m}{\delta}}{2(m - 1)}}.
\end{equation}
\end{theorem}

When the input data $\| x \| \le B$ and the Frobenius norms of the $D$ weight matrices are upper bounded by $M_F(1), \ldots, M_F(D)$, respectively, Neyshabur {\it et al.} \cite{neyshabur2015norm} prove via Rademacher complexity that with high probability, the generalization error scales as
\begin{equation}
\label{eq:margin_bound_neyshabur2015norm}
\mathcal O\left( \frac{B2^D\prod_{j = 1}^DM_F(j)}{\sqrt m} \right).
\end{equation}
However, this bound is exponentially dependent with the depth.

Then, through covering number, Dudley entropy integral, and Rademacher complexity, Bartlett {\it et al.} \cite{bartlett2017spectrally} obtain the following spectrally-normalized upper bound for the generalization error,
\begin{equation*}
\tilde{\mathcal O} \left( \frac{\|X\|_2  \log W}{\gamma m} R_{\mathcal A} + \sqrt{\frac{1/\delta}{m}}  \right),
\end{equation*}
where
\begin{equation*}
R_{\mathcal A} = \left( \prod_{i = 1}^D \rho_i \| A \|_\sigma  \right) \left( \sum_{i = 1}^D \frac{\|A_i^\top - M_i^\top\|_{2, 1}^{2/3}}{ \| A \|_\sigma^{2/3}}  \right)^{3/2}.
\end{equation*}
This bound is strictly tighter than Neyshabur {\it et al.} \cite{neyshabur2015norm}. Moreover, this bound suggests that one can boost the generalizability via controlling the spectral norm of the network. This explains spectral normalization \cite{miyato2018spectral}, an effective regularizer mainly used in generative adversarial networks, and singular value bounding \cite{jia2017improving} for boosting neural network training performance.

Concurrently, Neyshabur {\it et al.} \cite{neyshabur2017pac} prove a similar generalization bound as below via the PAC-Bayesian theory,
\begin{equation*}
\mathcal O \left( \sqrt{\frac{B^2 D^2 W \log (DW) \prod_{i = 1}^D \frac{\|W_i\|_F^2}{\|W_i\|_2^2} + \log \frac{D m}{\delta}}{\gamma^2 m}} \right).
\end{equation*}
The proofs are in two-step: (1) they prove a perturbation bound which controls the changes of the output when perturbation introduced to the weights. Thereby, we can bound the sharpness of the network; and (2) the perturbation bound further leads to a generalization bound via the PAC-Bayesian theory. However, this bound has been shown strictly weaker than the one given by Bartlett {\it et al.} \cite{bartlett2017spectrally}.

Through introducing some norm constraints to the weights, Golowich {\it et al.} \cite{golowich2017size} improve the exponential depth dependence of eq. (\ref{eq:margin_bound_neyshabur2015norm}) by Neyshabur {\it et al.} \cite{neyshabur2015norm} to polynomial dependence as follows,
\begin{equation*}
\mathcal O\left( \frac{B \sqrt D\prod_{j = 1}^DM_F(j)}{\sqrt m} \right),
\end{equation*}
or
\begin{equation*}
\mathcal O\left( \frac{B \sqrt{D+\log m} \prod_{j = 1}^DM_F(j)}{\sqrt m} \right).
\end{equation*}
When assuming
\begin{equation*}
\prod_{j = 1}^D \| W_j \|^2 \le M,
\end{equation*}
where $M > 0$ is a constant and $W_j$ is the $j$-th weight matrix, Golowich {\it et al.} \cite{golowich2017size} further improves the bounds to depth-independent as follows,
\begin{equation*}
\tilde{\mathcal O} \left( B \left( \prod_{j = 1}^D M_F(j)\right) R_{G} \right),
\end{equation*}
where 
\begin{equation*}
R_G = \min \left\{ \sqrt{\frac{\log \left( \frac{1}{\Gamma}  \prod_{j = 1}^DM_F(j) \right)}{\sqrt{m}}}, \sqrt{\frac{D}{m}} \right \},
\end{equation*}
and $\Gamma$ is the upper bound of the spectral norm of all the weight matrices.

Moreover, Golowich {\it et al.} \cite{golowich2017size} also prove a lower bound as follows,
\begin{equation*}
\Omega \left( \frac{B \prod_{j = 1}^D M_F(j) W^{\max\{ 0, \frac{1}{2} - \frac{1}{p} \}}}{\sqrt{m}} \right).
\end{equation*}
It suggests that a tight bound would be $\mathcal O\left( \frac{1}{\sqrt{m}} \right)$.

\section{Stochastic gradient descent: An implicit regularization}
\label{sec:SGM}

The over-parameterization makes a colossal hypothesis space. Despite that Golowich {\it et al.} \cite{golowich2017size} have provided generalization bounds explicitly independent with the network size, one might observe some implicit dependence. For example, the factor $M_F(j), j = 1, \ldots, D,$ are the upper bounds of weight norms which might be also influenced by the model size.

Stochastic gradient methods (SGM), including stochastic gradient descent (SGD) and its variants, are the dominant optimization methods in deep learning. They only explore a small part of the hypothesis space. This performs an implicit regularization to the deep learning models that controls the effective model capacity in an algorithm-dependent and data-dependent manner. Therefore, approaching algorithm-dependent and data-dependent bounds via SGM may overcome the conventional statistical learning theory based on VC dimension or Rademacher complexity.

\subsection{Stochastic gradient methods}

To optimize the expected risk, a natural tool is the gradient descent (GD). Specifically, the gradient of the expected risk in terms of the parameter $\theta$ below,
\begin{align*}
	g(\theta(t)) \triangleq & \nabla_{\theta(t)} \mathcal R(\theta(t)) 
	=  \nabla_{\theta(t)} \mathbb E_{(X, Y)} l(F_{\theta(t)}(X), Y),
\end{align*}
and the corresponding update equation are defined as follows,
\begin{gather*}
	\theta(t + 1) = \theta(t) - \eta g(\theta(t)),
\end{gather*}
where $\theta(t)$ is the parameter at the interation $t$ and $\eta > 0$ is the learning rate.

Stochastic gradient descent (SGD) estimates the gradient from mini batches of the training sample set to estimate the gradient $g(\theta)$. Let $S$ be the indices of a mini batch, in which all indices are independently and identically (i.i.d.) drawn from $\{ 1, 2, \ldots, N \}$, where $N$ is the training sample size. Then similar to the gradient, the iteration of SGD on the mini-batch $S$ is defined as follows, 
\begin{align}
\label{eq:stochastic_gradient}
	\hat{g}_{S} (\theta(t)) = & \nabla_{\theta(t)} \hat{\mathcal R}(\theta(t)) 
	=  \frac{1}{|S|} \sum_{n \in S} \nabla_{\theta(t)} l(F_{\theta(t)}(X_{n}),
	Y_{n}),
\end{align}
and
\begin{gather}
\label{eq:stochastic_gradient_descent}
	\theta(t + 1) = \theta(t) - \eta \hat g(\theta(t)),
\end{gather}
where
\begin{equation*}
\hat{\mathcal R}(\theta) = \frac{1}{|S|} \sum_{n \in S} l(F_{\theta}(X_{n}), Y_{n})
\end{equation*}
is the empirical risk on the mini batch and $|S|$ is the cardinality of the set $S$. For brevity, we rewrite
\begin{equation*}
l(F_{\theta}(X_{n}), Y_{n}) = l_{n}(\theta)
\end{equation*}
in the rest of this paper.

Also, suppose that in step $i$, the distribution of parameter is $Q_{i}$, the initial distribution is $Q_{0}$, and the convergent distribution is $Q$. Then SGD is used to find $Q$ from $Q_{0}$ through a series of $Q_{i}$.



\subsection{Generalization bounds on convex loss surfaces}

When the loss is continuous and convex, the generalizability of SGMs has been intensively studied.

Ying and Pontil \cite{ying2008online} prove the generalization bound in the following rate, when the step size is fixed: $\eta_t \simeq m^{-\frac{2\zeta}{2\zeta + 1}}$,
\begin{equation*}
\mathcal O\left( m^{-\frac{2\zeta}{2\zeta + 1}} \log m \right),
\end{equation*}
where $\zeta$ is a constant.

Then, Dieuleveut {\it et al.} \cite{dieuleveut2016nonparametric} improves the result to the following rate for the average, when $2 \zeta + \gamma > 1$,
\begin{equation*}
\mathcal O\left( m^{-\frac{2 \min \{\zeta, 1\}}{2 \min \{\zeta, 1\} + \gamma}} \right).
\end{equation*}

Other related works include Lin {\it et al.} \cite{lin2016generalization}; Lin and Rosasco \cite{lin2016optimal}; Chen {\it et al.} \cite{chen2016statistical}; Wei {\it et al.} \cite{wei2017early}. However, the convex assumption for the loss surface is usually too strong in deep learning. However, the loss surface is usually highly non-convex in deep learning. This makes the aforementioned results invalid.

\subsection{Generalization bounds on non-convex loss surface}

Results on non-convex loss surface have also been seen in the literature. Related works include generalization bounds via algorithmic stability and PAC-Bayesian theory. 

\textbf{Algorithmic stability.} Bousquet and Elisseeff \cite{bousquet2002stability} propose to employ {\it algorithmic stability} to measure the stability of the output hypothesis when the training sample set is disturbed. Algorithmic stability has many versions. A popular one is as follows.

\begin{definition}[uniform stability; cf. \cite{bousquet2002stability}, Definition 6]
\label{def:uniform_stability}
A machine learning algorithm $\mathcal A$ is uniformly stable, if for any neighboring sample pair $S$ and $S'$ which are different by only one example, we have the following inequality,
\begin{equation*}
  \left\vert \mathbb E_{\mathcal{A}(S)}l(\mathcal{A}(S), Z) - \mathbb E_{\mathcal{A}(S')}l(\mathcal{A}(S'), Z) \right\vert \le \beta,
\end{equation*}
where $Z$ is an arbitrary example, $\mathcal{A}(S)$ and $\mathcal{A}(S')$ are the output hypotheses learned on the training sets $S$ and $S'$, respectively, and $\beta$ is a positive real constant. The constant $\beta$ is called the uniform stability of the algorithm $\mathcal A$.
\end{definition}

It is natural that an algorithm insensitive to the disturbance the training data have good generalization abilities. Following this intuition, some generalization bounds have been proved based on the algorithmic stability \cite{bousquet2002stability, xu2011sparse}. For example, based on the uniform stability, Bousquet and Elisseeff \cite{bousquet2002stability} prove the following theorem for the following {\it generalization in expectation}.

\begin{theorem}
Let algorithm $\mathcal A$ is $\beta$-uniformly stable. Then,
\begin{equation*}
|\mathbb E_{S, \mathcal A} [\hat{\mathcal R}[\mathcal A(S)] - \mathcal R[\mathcal A(S)]]| \le \beta.
\end{equation*}
\end{theorem}


Recent works usually model SGM via stochastic differential equations (SDEs). Mini-batch stochastic gradients introduce randomness into SGM ({\it i.e.}, the stochastic gradient can be decomposed into sample-estimation of the gradient on the whole training set plus a noise). So the weight update in SGM can be formulated as a stochastic process. The output hypothesis is then drawn from the steady distribution of this stochastic process. We can eventually analyze the generalization and optimization of deep learning via the steady distribution.

SGM is usually formulated by eqs. (\ref{eq:stochastic_gradient}) and (\ref{eq:stochastic_gradient_descent}). The stochastic gradient introduces gradient noise into the weight updates. When the gradient noise is modeled by a Gaussian distribution, SGM reduces to stochastic gradient Langevin dynamics(SGLD) as follows \cite{mandt2017stochastic},
\begin{align}
\label{eq:OU_process}
	\Delta \theta(t) = & \theta(t + 1) - \theta(t) 
	= - \eta \hat{g}_{S}(\theta(t)) 
	= - \eta g(\theta) + \sqrt{\frac{2\eta}{\beta}} \Delta W \nonumber, \Delta W \sim N(0, I),
\end{align}
where $\eta$ is the learning rate and $\beta > 0$ is the inverse temperature parameter.

Hardt {\it et al.} \cite{hardt2016train} prove the following upper bound for uniform stability, which exponentially depends on aggregated step sizes and smoothness parameter. 

\begin{theorem}
Suppose the loss function is $\epsilon$-smooth, $L$-Lipschitz, and not larger than $1$ for every data. We run SGM with monotonically increasing step sizes $\alpha_t \le c/t$ for $T$ steps. Then, SGM is uniform stable with
\begin{equation*}
\beta \le \frac{1 + 1/\beta c}{m - 1} (2cL^2)^{\frac{1}{\beta c + 1}} T^{\frac{1}{\beta c + 1}}.
\end{equation*}
\end{theorem}

This proof of this theorem mimics the proofs for convergence. Suppose the loss function $l$ is $L$-Lipschitz constant with respect to the weight for any example; {\it i.e.},
\begin{equation*}
\mathbb |l(w; z) - l(w'; z)| \le L \|w - w'\|.
\end{equation*}
In other words, the stabilities can be measured by the stability of the weight. Therefore, one can just analyze how the weight diverges as a function of time $t$. The authors then deliver generalization bounds relying on the step sizes and the iteration numbers. Further, we can easily obtain a generalization bound based on this theorem.

Under five more assumptions, Raginsky {\it et al.} \cite{raginsky2017non} prove the following upper bound for the expected excess risk,
\begin{align*}
& \tilde{\mathcal O} \left( \frac{\beta ( \beta + U )^2}{\lambda_*} \delta^{1/4} \log \left( \frac{1}{\epsilon} + \epsilon \right) \right) 
+ \tilde{\mathcal O} \left( \frac{(\beta + U)^2}{\lambda_* + m} + \frac{U \log(\beta + 1)}{\beta} \right),
\end{align*}
provided,
\begin{equation*}
k = \tilde{\Omega} \left( \frac{\beta ( \beta + U )}{\lambda_*\epsilon^4} \log^5  \left( \frac{1}{\epsilon} \right) \right),
\end{equation*}
and
\begin{equation*}
\eta \le \left( \frac{\epsilon}{\log(1/\epsilon)} \right)^4,
\end{equation*}
where $U$ is the parameter size, $m$ is the training sample size, the inverse temperature parameter $\beta > 1 \vee 2/m$, $\lambda_*$ is a spectral gap parameter that controls the exponential rate of convergence of the dynamic of SGM to its stationary distribution, $d$ is the factor concerning the dissipativity of the hypothesis $h$; {\it {\it i.e.}}, for any datum $z$ and any parameter $w$, for some $m > 0$ and $b > 0$,
\begin{equation*}
<w, \nabla h(w, z)> \le m \| w \|^2 - b,
\end{equation*}
and 
\begin{equation*}
\epsilon \in \left(0, \frac{m}{4M^2} \wedge e^{-\tilde{\Omega} \left( \frac{\lambda_*}{\beta ( \beta + U )} \right)}\right).
\end{equation*}
The proofs are completed in the following three steps: (1) when the step size is small enough, the authors prove that the dynamics of SGLD converge to continuous-time Langevin dynamic processes under $2$-Wasserstein distance; (2) the authors then show that the Langevin dynamic then converge to a Gibbs distribution when the training time has been sufficiently long; and (3) the authors construct a Gibbs algorithm where the output hypothesis is almost an empirical risk minimizer. Also, the Gibbs algorithm is proven to be algorithmic stable under $2$-Wasserstein distance. However, this result is based on the convergence of the weight distribution to the stationary distribution, which usually exponentially relies on the model dimension. 

A canonical mutual information-based generalization bound is given in Xu and Raginsky \cite{xu2017information},
\begin{equation}
\label{eq:xu2017information}
\sqrt{\frac{2\sigma^2}{m}I(S; \theta)},
\end{equation}
where $S$ is the training sample, $\theta$ is the model parameter, and the loss is $\sigma$-sub-Gaussian under the data distribution. This result stems from the following result on the informational theoretical ``stability'' of a two-input function $f(X, Y)$,
\begin{equation*}
|f(X, Y) - f(\bar X, \bar Y)| \le \sqrt{2 \sigma^2 I(X; Y)},
\end{equation*}
if the $f(\bar X, \bar Y)$ is sub-Gaussian under distribution $P_{\bar X, \bar Y} = P_X \otimes P_Y$ where $P_X$ and $P_Y$ are distributions of $X$ and $Y$, respectively. It is worth noting that the mutual information is calculated between the learned weights and all training data; it is an ``on-average'' estimate of how much information has been transformed into the learned model from the data. 

Based on this result, Pensia {\it et al.} \cite{pensia2018generalization} gives a generalization error bound for SGLD,
\begin{equation}
\label{eq:pensia2018generalization}
\sqrt{\frac{R^2}{m} \sum_{t = 1}^T \frac{\eta_t^2 L^2}{\sigma^2_t}},
\end{equation}
where the hypothesis is assumed to $R$-sub-Gaussian, $\eta_t$ is the learning rate in the $t$-th iteration, $L$ is the upper bound of the weight update $\Delta \theta_t$, and the gradient noise is assumed to be $N(0, \sigma_t^2 I)$. This result is then extended by Bu {\it et al.} \cite{bu2020tightening}:
\begin{itemize}
\item
Assume the loss $l(\theta, z)$ is $R$-sub-Gaussian under $z \sim \mu$, where $z_i$ is the $i$-th datum, $\theta$ is the model parameter, and $\mu$ is the data generating distribution. Then, the generalization error has the following upper bound,
\begin{equation}
\label{eq:bu_bound}
\frac{1}{m} \sum_{i = 1}^m \sqrt{2R^2 I(\theta; z_i)},
\end{equation}
where $m$ is the sample size.

\item
Assume the loss $l(\tilde \theta, \tilde z)$ is $R$-sub-Gaussian under distribution
\begin{equation*}
P_{\tilde \theta, \tilde z} = P_\theta \otimes P_z.
\end{equation*}
Then, the generalization error has the same upper bound (eq. \ref{eq:bu_bound}).
\end{itemize}
In contrast with Xu and Raginsky \cite{xu2017information}, Bu {\it et al.} \cite{bu2020tightening} proposes an ``individual'' version metric to measure the information transformed into the learned model from the data; see eq. (\ref{eq:bu_bound}) where the whole training sample $S$ is replaced by individual datums $I(\theta; z_i)$.

Mou {\it et al.} \cite{mou2017generalization} models SGLD as a Langevin diffusion dynamic and then prove an $\mathcal O\left(1/m\right)$ generalization bound and an $\mathcal O\left(1/\sqrt{m}\right)$ generalization bound for SGLD via the stability and the PAC-Bayesian theory, respectively:
\begin{itemize}
\item
\textbf{Algorithmic stability:} Assume that the loss function is upper bounded by $C$, the hypothesis is $L$-Lipschitz continuous, and the step size 
$\eta_t \le \frac{\log 2}{\beta L^2}$ 
in any iteration $t$, the generalization error has the following upper bound in average, 
\begin{equation}
\label{eq:mou2017generalization_1}
\mathcal O \left( \frac{2LC}{m} \left( \beta \sum_{t = 1}^N \eta_t \right)^2 \right).
\end{equation}

\item
\textbf{PAC-Bayesian theory:} Assume the empirical risk is regularized by the $l_2$ norm of the weight; {\it {\it i.e.}}, $\frac{\lambda}{2} \|\theta\|^2$, and the loss function is sub-Gaussian. Then, we have the following high-probability generalization bound,
\begin{equation*}
\mathcal O \left( \sqrt{\frac{\beta}{m} \sum_{k = 1}^N \eta_k e^{-\frac{\lambda}{3}(T_N - T_k)} \mathbb E[\| g_k(\theta_k) \|^2]} \right),
\end{equation*}
where $N$ is the iteration number, 
$T_k = \sum_{i = 1}^k \eta_i$, 
$\beta$ is the temporal parameter, $g_k(\theta_k)$ is the gradient in the $k$ iteration, 
\end{itemize}

He and Su \cite{he2020local} empirically shows that neural networks are {\it locally elastic}: the prediction on an instance $x'$ is not sensitive to learning on a dissimilar example $x$ by SGD. This phenomenon inspires a localized version of algorithmic stability by Deng {\it et al.} \cite{deng2020toward}.

\begin{definition}[locally elastic stability; cf. Deng {\it et al.} \cite{deng2020toward}, Definition 1]
\label{def:locally_elastic_stability}
A machine learning algorithm $\mathcal A$ has locally elastic stability $\beta_m(\cdot, \cdot)$ with respect to the loss function $l$, if for any sample $S \in \mathcal Z^m$, $z_i \in S$, and $z \in \mathcal Z$, we have the following inequality,
\begin{equation*}
\left\vert l(\mathcal{A}(S), z) - l(\mathcal{A}(S^{-i}), z) \right\vert \le \beta_m(z_i, z).
\end{equation*}
\end{definition}

This localized algorithmic stability leads to a tighter generalization bound.

\begin{theorem}
Let algorithm $\mathcal A$ has $\beta_m(\cdot, \cdot)$-locally elastic stability. Assume that $\| z \| \le 1$ for any $z \in \mathcal Z$ and $\beta_m(\cdot, \cdot) = \frac{\beta(\cdot, \cdot)}{m}$ where $\beta(\cdot, \cdot)$ is independent with the sample size $m$ and $\|\beta(\cdot, \cdot)\| \le M_\beta$. Then, for any $0 < \delta < 1$ and $\eta > 0$, with probability at least $1 - \delta$, we have
\begin{equation*}
\left|\mathcal R[\mathcal A(S)] - \hat{\mathcal R}[\mathcal A(S)]\right| \le \frac{2 \sup_{z' \in \mathcal Z} \mathbb E_z \beta(z', z)}{m} + 2 \left( 2 \sup_{z' \in \mathcal Z} \mathbb E_z \beta(z', z) + \eta + M_l \right) \sqrt{\frac{2 \log(2/\delta)}{m}}.
\end{equation*}
\end{theorem}

Deng {\it et al.} then extend the definition of locally elastic stability to stochastic algorithms, including SGD. The authors prove that SGD is locally elastic stable under the extended definition. This helps approach characterizing the effective hypothesis space that the SGD explores.

Moreover,  Chen {\it et al.} \cite{chen2018stability} proves a trade-off between convergence and stability for all iterative algorithms respectively under convex smooth setting and strong convex smooth setting, which leads to an $\mathcal O\left(1/m\right)$ generalization bound. Other advances include the works by He {\it et al.} \cite{he2019control}, Tzen \cite{tzen2018local}, Zhou \cite{zhou2018generalization}, Wen {\it et al.} \cite{wen2019interplay}, and Li {\it et al.} \cite{li2019generalization}.

\subsection{Generalization bounds relying on data-dependent priors}

In Bayesian statistics, priors are given according to the ``practice experience'' or domain knowledge. In algorithm designing, one usually cannot assume they have understood the data. Thus, the prior needs to be independent with the training sample. Bayesian statistics usually adopts priors containing no ``information'' of the data, such as uniform distributions or Gaussian distributions. Moreover, the impact of prior usually goes to vanishing when training progress. From this aspect, setting data-independent priors would not comprise the performance in an asymptotic manner. Recent works slightly relax the restrictions in convention and advance the PAC-Bayesian theory. 

\textbf{Distribution-dependent priors.} Some works design priors relying on the data generation distribution but still not directly relying on the training data. This would be reasonable since we can assume the data distribution has been fixed before the data was collected \cite{lever2013tighter}. Such {\it distribution-dependent} priors have shown to be able to considerably tighten the generalization bounds.

Following this line, Lever {\it et al.} \cite{lever2013tighter} tightens the PAC-Bayesian bound (see Theorem \ref{thm:PAC-Bayesian}) to the following bound. For any positive constant $C$, we have that at least probability $1 - \delta > 0$ with respect to the drawn sample, the generalization error is upper bounded as follows,
\begin{align*}
& \mathcal R(Q(h)) - C^* \hat{\mathcal R}(Q(h)) 
\le \frac{C^*}{Cm} \left( \lambda \sqrt{\frac{2}{m} \log\frac{2\xi(m)}{\delta}} + \frac{\gamma^2}{2m} + \log \frac{2}{\delta} \right),
\end{align*}
where $Q(h)$ is the posterior of the learned model, its density is as below,
\begin{equation*}
q(h) \propto e^{- \gamma \hat{\mathcal R}(Q(h)},~
\xi(m) = \mathcal O(\sqrt{m}),~
C^* = \frac{C}{1 - e^{-C}},
\end{equation*}
and
$\gamma$ is a positive constant.

Li {\it et al.} \cite{li2019generalization} further design a {\it Bayes-stability} framework to deliver generalization bounds based on distribution-dependent priors. They tighten the result of Mou {\it et al.} \cite{mou2017generalization} (eq. \ref{eq:mou2017generalization_1}) to the follows,
\begin{equation*}
\mathcal O \left( \frac{C}{m} \sqrt{\mathbb E_S \left[ \sum_{t=1}^T \frac{\eta^2_t}{\sigma_t^2} g_e(t) \right]} \right),
\end{equation*} 
where the gradient noise is assumed to be $\frac{\sigma_t}{\sqrt{2}} I$, $C > 0$ is the upper bound of the loss function, $T$ is the number of iterations, $\eta_t$ is the learning rate in the $t$-th iteration, and
\begin{equation*}
g_e(t) = \mathbb E_{W_{t-1}} \left[ \frac{1}{m} \sum_{i=1}^m \| \nabla F(W_{t-1}, z_i) \|^2 \right].
\end{equation*}

\textbf{Data-dependent priors.} Negrea {\it et al.} \cite{negrea2019information} further push the frontier that constructs priors not independent with data. Suppose $S_j$ is a subset of $\subset S$ with size of $n < m$. One may design a prior exploiting $S_J$ to deliver a data-dependent forecast of the posterior $Q$. We denote the subset as $S_J=\{ z_{j_1}, \ldots, z_{j_n} \}$, where all indices constitute a set $J$. The index set $J$ is randomly drawn from $\{1, \ldots, m\}$. Suppose that we have the following generalization bound,
\begin{equation*}
\mathbb E^{\mathcal F} \left[ \mathcal R(\theta) - \hat{\mathcal R}_{S_J}(\theta) \right] \le B,
\end{equation*}
where $\hat{\mathcal R}_{S_J}$ is the empirical risk on the set $S_J$, $B > 0$ is the bound, $\mathcal F$ is a $\sigma$-field such that $\sigma(S_J) \subset \mathcal F \bot \sigma(S_J^c)$. Also, we denote by $U$ the uncertainty introduced by the machine learning algorithm. Negrea {\it et al.} \cite{negrea2019information} prove the following theorem, which considerably improve the one given in Xu and Raginsky \cite{xu2017information} (eq. \ref{eq:xu2017information}),

\begin{theorem}
Let $J \subset [m]$, $|J| = n$ be uniformly distributed and independent from the training sample set $S$ and the model parameter $\theta$. Suppose the loss is $\sigma$-sub-Gaussian under the data distribution. When the posterior $Q = \mathbb P^S(\theta)$ and the prior $P$ be a $\sigma(S_J)$-measurable data-dependent on the data space,
\begin{align*}
\mathbb E \left[ \mathcal R(\theta) - \hat{\mathcal R}(\theta) \right] 
\le & \sqrt{\frac{2 \sigma^2}{m - n} I(\theta; S_J^c)} \nonumber\\
\le & \sqrt{\frac{2 \sigma^2}{m - n} \mathbb E [\text{KL}(Q||P)]}.
\end{align*}
Moreover, let $U$ be independent with $S$ and $\theta$. When the posterior $Q = \mathbb P^{S, U} (\theta)$ and the prior $P$ be a $\sigma(S_J, U)$-measurable data-dependent on the data space,
\begin{align*}
\mathbb E \left[ \mathcal R(\theta) - \hat{\mathcal R}(\theta) \right] 
\le & \sqrt{\frac{2 \sigma^2}{m - n} I^{S_J, U}(\theta; S_J^c)} \nonumber\\
\le & \sqrt{\frac{2 \sigma^2}{m - n} \mathbb E^{S_J, U} [\text{KL}(Q||P)]}.
\end{align*}
\end{theorem}

Following this path, further works include Haghifam {\it et al.} \cite{haghifam2020sharpened} cooperating with Steinke and Zakynthinou \cite{steinke2020reasoning}, Dziugaite {\it et al.} \cite{dziugaite2020role}, Hellström and Durisi \cite{hellstrom2020generalization}, and their applications \cite{cherian2020efficient}.

\subsection{The role of learning rate and batch size in shaping generalizability} 

\begin{figure*}[t]
  \centering
  \includegraphics[width=\linewidth]{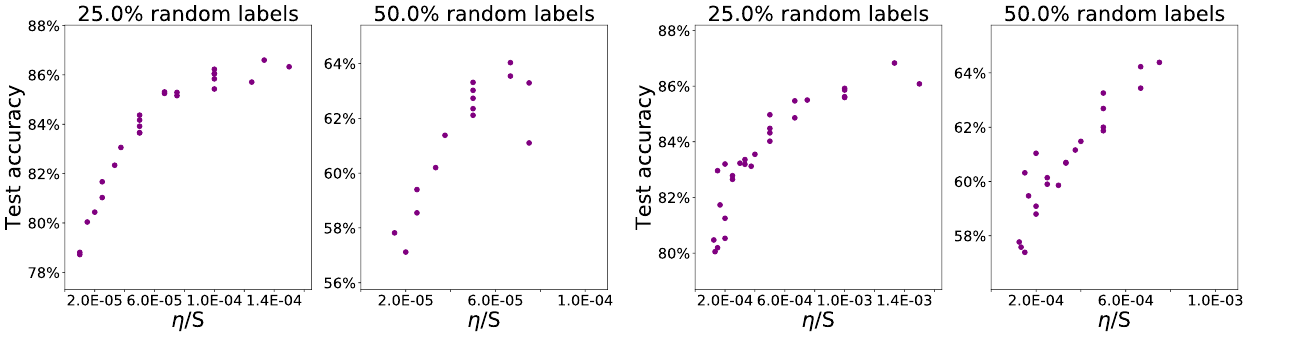}
  \caption{Scatter plots of test accuracy to ratio of batch size to learning rate. See \cite{jastrzkebski2017three}.}
  \label{fig:va_ratio_jastrzebski}
\end{figure*}

\begin{figure}[t]
    \centering
        \subfloat[Test accuracy-batch size\label{fig:va_bs}]{\includegraphics[width=0.45\linewidth]{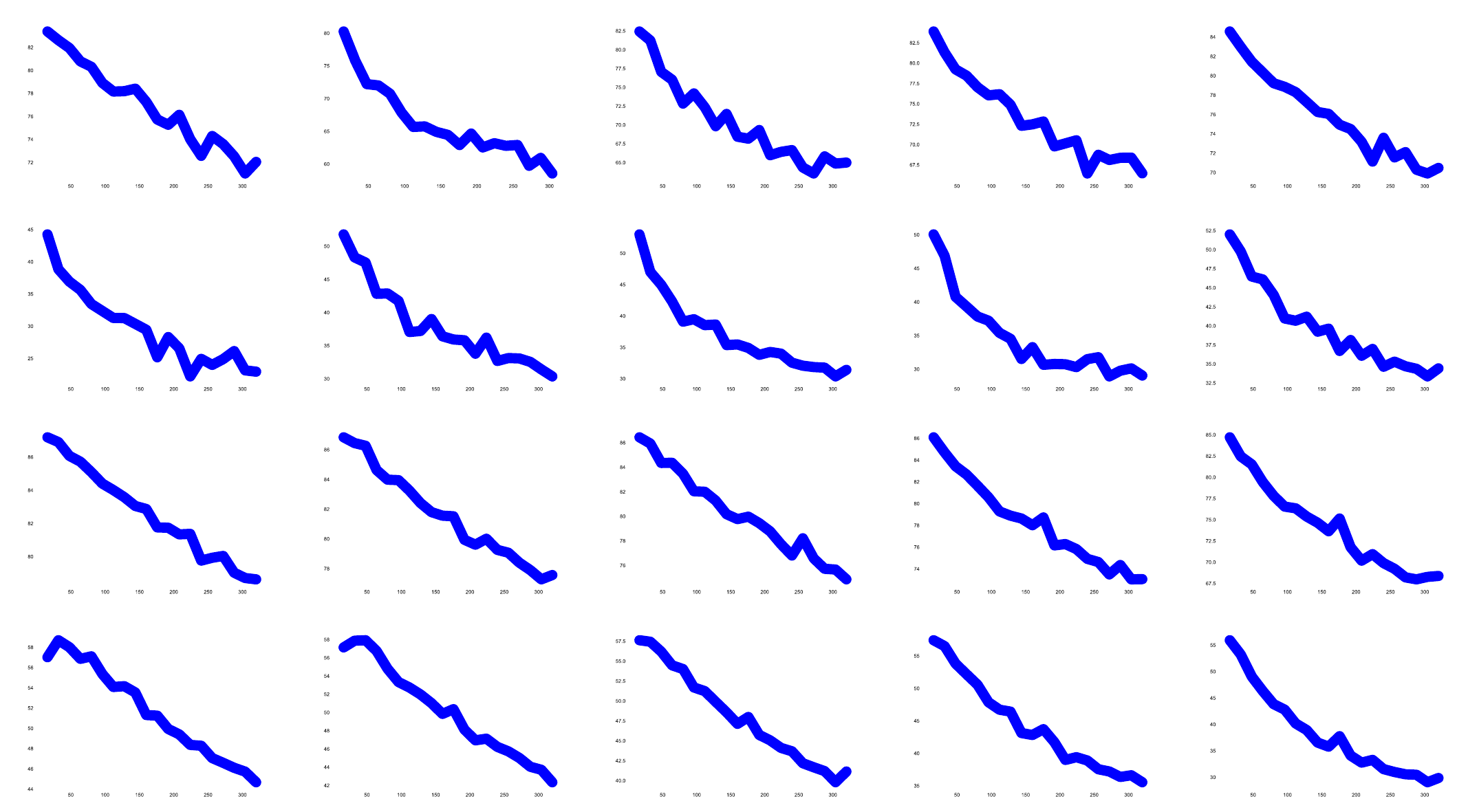}}
        \qquad
        \subfloat[Test accuracy-learning rate\label{fig:va_lr}]{\includegraphics[width=0.45\linewidth]{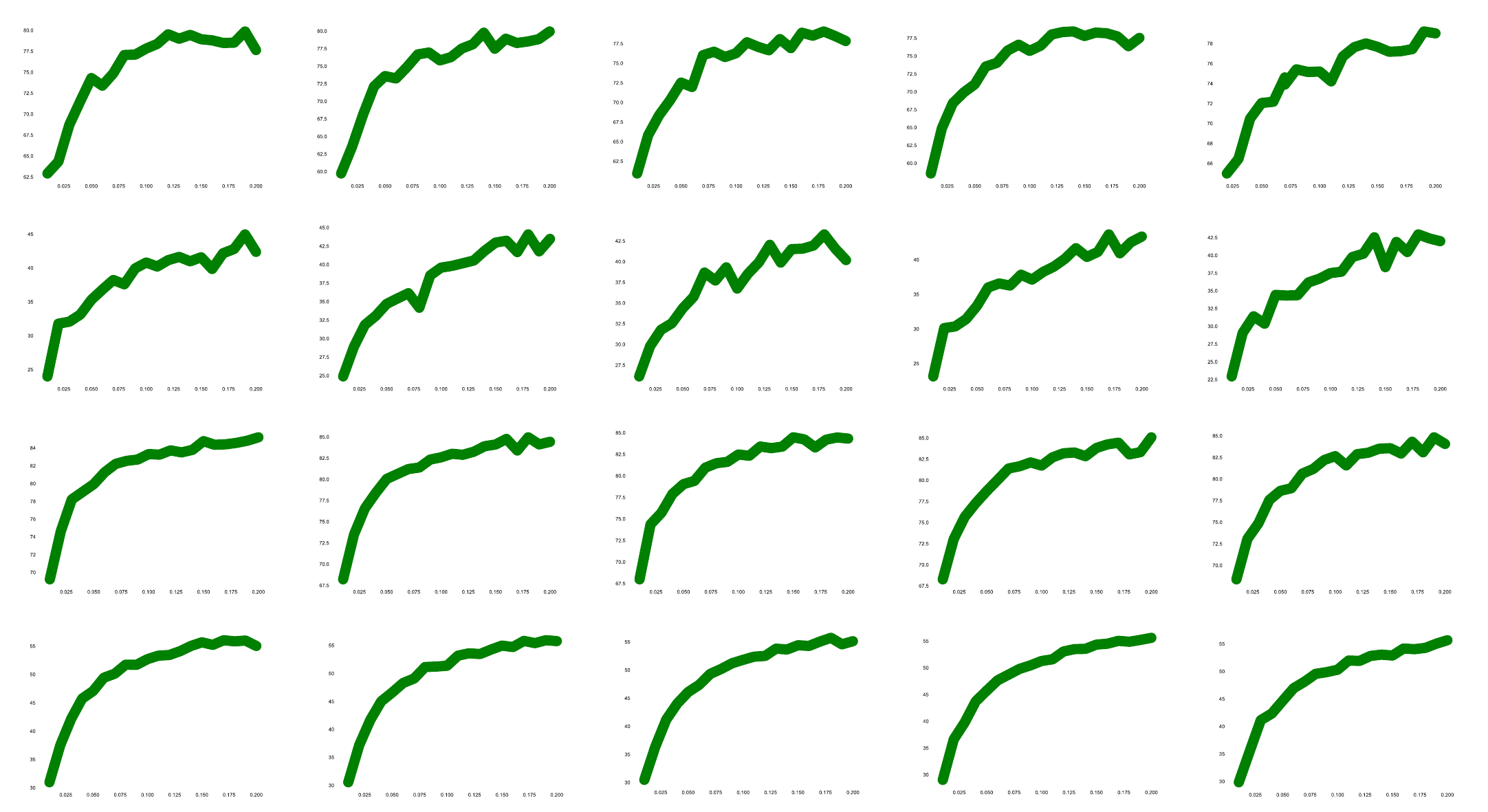}}
    \caption{Curves of test accuracy to batch size and learning rate. The four rows are respectively for (1) ResNet-110 trained on CIFAR-10, (2) ResNet-110 trained on CIFAR-100, (3) VGG-19 trained on CIFAR-10, and (4) VGG-19 trained on CIFAR-10. Each point represents a model. Each curve is based on 20 networks. See \cite{he2019control}.}
    \label{}
\end{figure}

\begin{figure}[t]
  \centering
  \includegraphics[width=\linewidth]{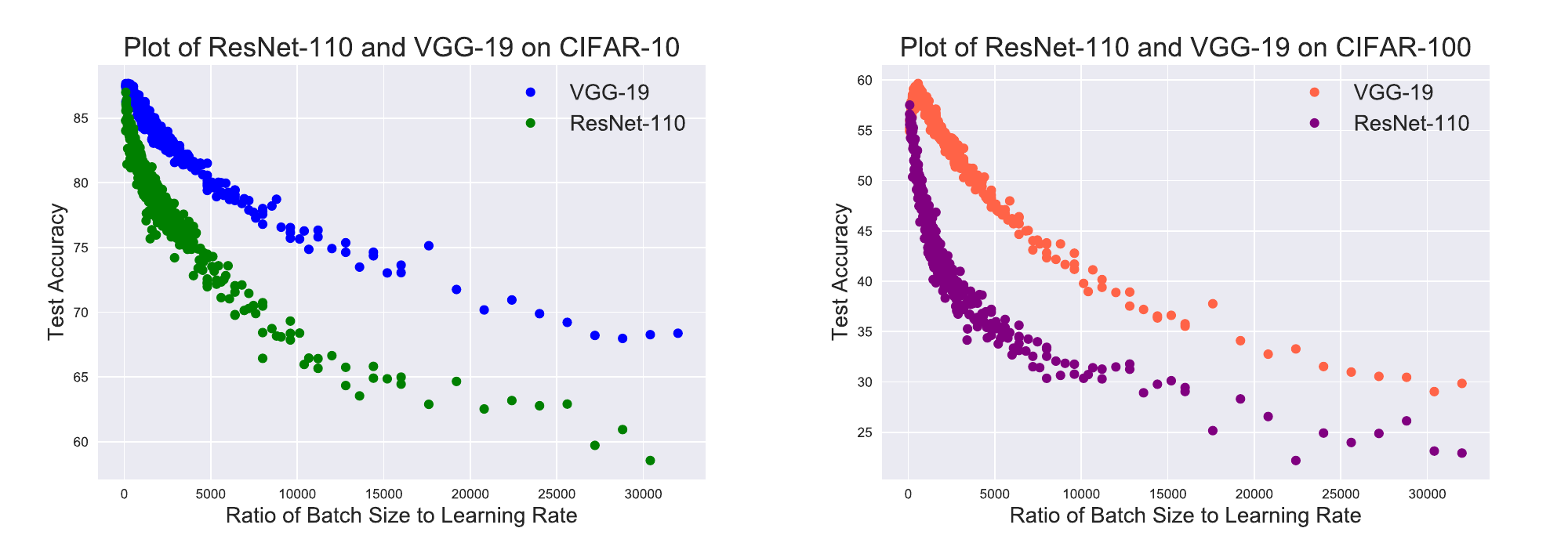}
  \caption{Scatter plots of accuracy on test set to ratio of batch size to learning rate. The four colors, blue, green, orange, and purple, are for results of (1) VGG-19 on CIFAR-10, (2) ResNet-110 on CIFAR-10, (3) VGG-19 on CIFAR-100, and (4) ResNet-110 on CIFAR-100. Each point represents a model. Totally 1,600 points are plotted. See \cite{he2019control}.}
  \label{fig:va_ratio}
\end{figure}

Keskar {\it et al.} \cite{keskar2016large} conducted experiments which show large-batch SGM usually finds sharp local minima, while its small-batch counterpart leads to flat minima. The authors also explain that flatter local minima can make SGM generalize better. However, the experiments wherein only involve two values of batch size, which considerably limits the generality of the results. Then, Dinh {\it et al.} \cite{dinh2017sharp} argue that most current notions for sharpness or flatness are ill-defined. The flatness/sharpness is still under intensive study.

Goyal {\it et al.} \cite{goyal2017accurate} further propose a {\it linear scaling rule} for SGM: ``when the minibatch size is multiplied by $k$, multiply the learning rate by $k$''. A similar statement is also given by Smith and Le \cite{smith2017don}. Jastrzebski {\it et al.} \cite{jastrzkebski2017three} present empirical results on the relationships between the ratio $|S|/\eta$ and the dynamics, geometry, and the generalization, but still, only two values of both the batch size and learning rate are investigated. 

Similar statements on the relationship between the generalizability and the hyperparameters are also presented in an empirical work by Chaudhari and Soatto \cite{chaudhari2018stochastic}. Fig. \ref{fig:va_ratio_jastrzebski} gives plots of the test accuracies of a multi-layer perceptron (MLP) trained on the dataset MNIST \cite{lecun1998gradient} in term of the ratio of the batch size to the learning rate respectively when $25.0\%$ and $50.0\%$ labels are randomly settled (see the original figure in \cite{jastrzkebski2017three}). These plots express the memorization ability\footnote{The concept of memorization is given by \cite{zhang2016understanding}. It is defined as the ability of a model to fit a dataset with random labels. If the training error can be very small even when the training labels are random, we argue the model is ``memorizing'' data, but not ``learn'' the data. Therefore, it is an important index expressing the generalization ability of a model.} 

He {\it et al.} \cite{he2019control} proves a PAC-Bayesian generalization bound under some assumptions. 
The generalization bound also suggests a positive correlation between the generalizability and the ratio of the learning rate and the batch size. He {\it et al.} \cite{he2019control} then conduct experiments that show a clear correlation between the generalizability and the ratio of batch size to learning rate; see fig. \ref{fig:va_ratio}.

For more details and advances, please see recent works by Arpit {\it et al.} \cite{arpit2017closer}, Krueger {\it et al.} \cite{krueger2017deep}, Carlini {\it et al.} \cite{carlini2018secret}, and Smith and Le \cite{smith2018bayesian}. 




\subsection{Interplay of optimization and Bayesian inference}

An interesting interplay is observed in the research of optimization and Bayesian inference \cite{Xiang:EECS-2020-145}: (1) Bayesian inference (or sampling) can be treated as optimization in the probability space ({\it e.g.}, in stochastic gradient Markov chain Monte Carlo) and can be scaled through optimization ({\it e.g.}, in variational inference); and (2) stochastic gradient Markov chain Monte Carlo also performs Bayesian approximation.

\textbf{Bayesian inference.} For parametric machine learning models, Bayesian learning aims to obtain the posterior of model parameters and then approach the best parameter. However, the analytic expression of the posterior is unknown in most real-world cases. To solve this problem, classical methods employ Markov chain Monte Carlo (MCMC) methods to infer the posterior, such as the Metropolis-Hastings algorithms \cite{hastings1970monte, geman1984stochastic} and hybrid Monte Carlo (HMC; \cite{duane1987hybrid}). Bayesian inference has been used in many areas, including topic model \cite{larochelle2012neural, zhang2020deep}, Bayesian neural network \cite{louizos2017multiplicative, roth2018bayesian}, and and generative models \cite{kingma2014auto, goodfellow2014generative, kobyzev2020normalizing}.

\textbf{Scaling Bayesian inference via optimization.} Bayesian inference would be prohibitively time-consuming on large-scale data. To address this issue,

(1) {\it Stochastic gradient Markov chain Monte Carlo} (SGMC; \cite{ma2015complete}) introduces stochastic grsestimates \cite{robbins1951stochastic} into MCMC. The family of SGMCMC algorithms includes stochastic gradient Langevin dynamics (SGLD; \cite{welling2011bayesian}), stochastic gradient Riemannian Langevin dynamics (SGRLD; \cite{patterson2013stochastic}), stochastic gradient Fisher scoring (SGFS; \cite{ahn2012bayesian}), stochastic gradient Hamiltonian Monte Carlo (SGHMC; \cite{chen2014stochastic}), stochastic gradient Nos\'e-Hoover Thermostat (SGNHT; \cite{ding2014bayesian}), etc. This paper analyses SGLD as an example of the SGMCMC scheme.

(2) {\it Variational inference} \cite{hoffman2013stochastic, blei2017variational, zhang2018advances} employs a two-step process to inference the posterior:
\begin{enumerate}
    \item We define a family of distributions,
    \begin{equation*}
    \mathcal{Q} = \{ q_\lambda | \lambda \in \Lambda\},
    \end{equation*}
    where $\lambda$ is the parameter.
    \item We search for the element from the distribution family $\mathcal{Q}$ that is the closest to the posterior $p(\theta|S)$, under KL divergence, {\it {\it i.e.}}, 
    \begin{equation*}
    \min_\lambda \mathrm{KL}(q_\lambda(\theta) \| p(\theta|S)).
    \end{equation*}
\end{enumerate}
Minimizing the KL divergence is equivalent to maximizing the evidence lower bound (ELBO) defined as below,
\begin{equation*}
\label{equ:elbo}
\text{ELBO}(\lambda,S) = \mathbb E_{q_\lambda} \log p(\theta, S) - \mathbb E_{q_\lambda} \log q_\lambda(\theta).
\end{equation*}

\textbf{Understanding stochastic gradient-based optimization via Bayesian inference.} E \cite{weinan2017proposal} proposed to understand the optimization in deep learning via stochastic differential equations. Mandt {\it et al.} 
\cite{mandt2017stochastic} propose to understand SGM as approximate Bayesian inference. Following this line, a large volume of literature has been delivered on SGM as shown in the previous subsections.

\section{Taking the high way: Study on loss surface}
\label{sec:loss}

A major barrier recognized by the whole community is that deep neural networks' loss surfaces are extremely non-convex and even non-smooth. Such non-convexity and non-smoothness make the analysis of the optimization and generalization properties prohibitively difficult. A natural idea is to bypass the geometrical properties and then approach a theoretical explanation. Some papers argue that such ``intimidating" geometrical properties are exactly the major factors that shape the properties of deep neural networks, and also the key to explaining deep learning.

\subsection{Does spurious local minimum exist?}

If all local minima are globally optimal, then we would not need to worry about optimization for deep learning. Then, is it true?

\subsubsection{Linear networks have no spurious local minima}

Linear neural networks refer to the neural networks whose activations are all linear functions. For linear neural networks, the loss surfaces do not have any spurious local minima: all local minima are equally good; they are all global minima.

Kawaguchi \cite{kawaguchi2016deep} proves that linear neural networks do not have any spurious local minimum under several assumptions: (1) the loss functions are squared losses; (2) $XX^T$ and $XY^T$ are both full-rank, where $X$ is the data matrix and $Y$ is the label matrix; (3) the dimension of the input layer is larger than the dimension of the output layer; (4) all the eigenvalues of matrix  $Y X^{\top}\left(X X^{T}\right)^{-1} X Y^{T}$ are different from each other. Lu and Kawaguchi \cite{lu2017depth} replace the assumptions to one: the data matrix $X$ and the label matrix $Y$ are both full-rank, which is, however, more restrictive.

Later, Zhou and Liang \cite{zhou2017critical} prove that all critical points are global minima when some conditions hold. The authors prove this based on a new result on the analytic formulation of the critical points.


\subsubsection{Nonlinear activations bring infinite spurious local minima}

However, this good property no longer stands when the activations are nonlinear. 

Zhou and Liang \cite{zhou2017critical} show that one-hidden layer ReLU neural networks have spurious local minima. However, the proof relies on that the hidden layer has only two nodes. Swirszcz {\it et al.} \cite{swirszcz2016local} relax the assumption to that most of the neurons are not activated. Safran and Shamir \cite{safran2017spurious} show that two-layer ReLU networks have spurious local minima in a computer-assisted. Yun {\it et al.} \cite{yun2018small} prove that one-hidden-layer ReLU neural networks have infinitely many spurious local minima when the outputs are one-dimensional. Goldblum {\it et al.} \cite{Goldblum2020Truth} prove that the training performance of all the local minima equals to a linear model for neural networks of any depth, and verify it by experiments. He {\it et al.} \cite{he2020piecewise} further extend the condition to arbitrary depth, arbitrary activations, and arbitrary output dimension.

\subsection{Geometric structure of loss surface.}

Then, how would the loss surface look like? Knowledge of the geometrical structures can considerably advance the current understanding of deep learning.

\textbf{Linear partition of the loss surface.} Soudry and Hoffer \cite{soudry2018exponentially} first highlight a {\it smooth and multilinear partition} exists on the loss surfaces of neural networks. The nonlinearities in the piecewise linear activations divide the loss surface into multiple smooth and multilinear open regions. Specifically, every nonlinear point in the activation functions creates a group of the non-differentiable boundaries between the cells, while the linear parts of activations correspond to the smooth and multilinear interiors.

He {\it et al.} \cite{he2020piecewise} prove several sophisticated properties of the partition: (1) if there are local minima within an open cell, they are all equally good in the sense of empirical risk, and therefore, all local minima are global minima within this cell; (2) the local minima in any open cell form an equivalence class. Moreover, these local minima are concentrated in a valley; and (3) when all activations are linear, the partition collapses to one single cell, covering linear neural networks as a special case.

\begin{figure}[t]
    \centering
        \subfloat[Spectrum of train Hessian on MNIST\label{fig:ResNet-56_no}]{\includegraphics[width=0.45\linewidth]{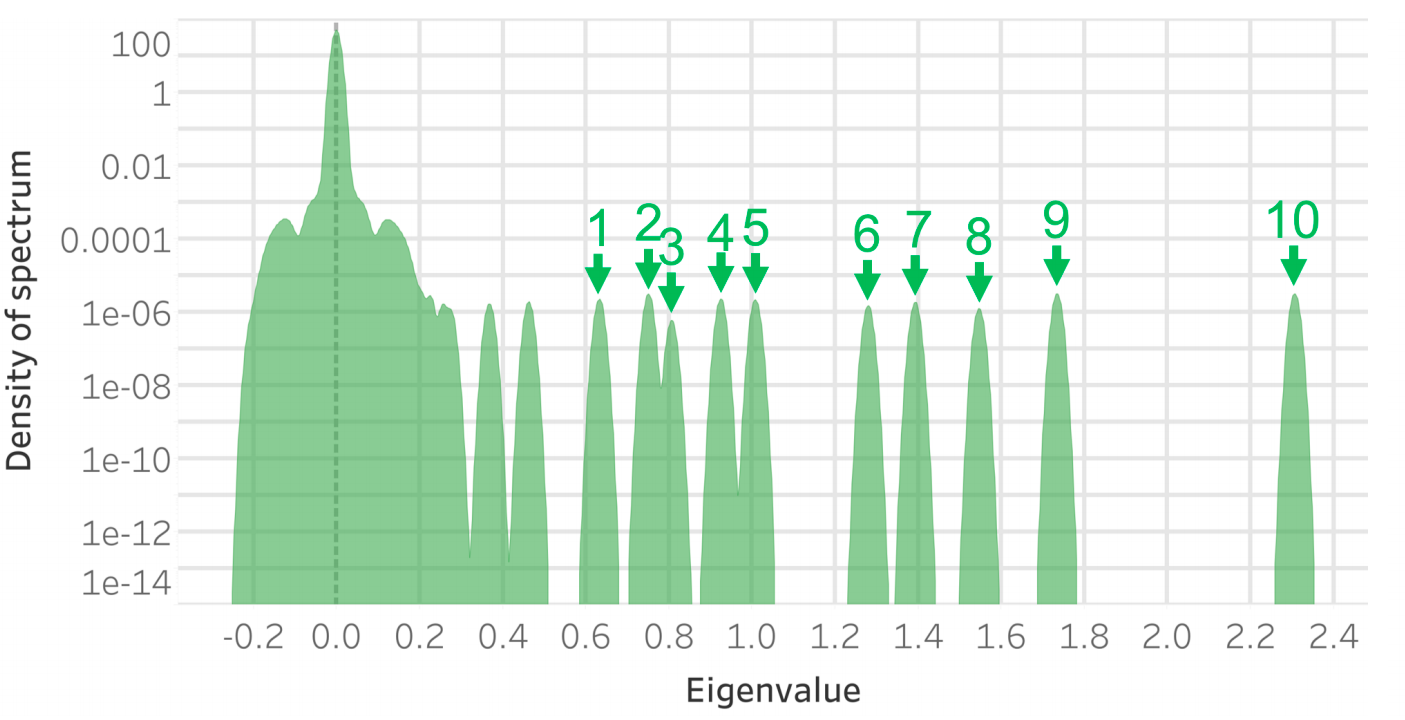}}
        \qquad
        \subfloat[Spectrum of train Hessian on CIFAR-10\label{fig:ResNet56}]{\includegraphics[width=0.45\linewidth]{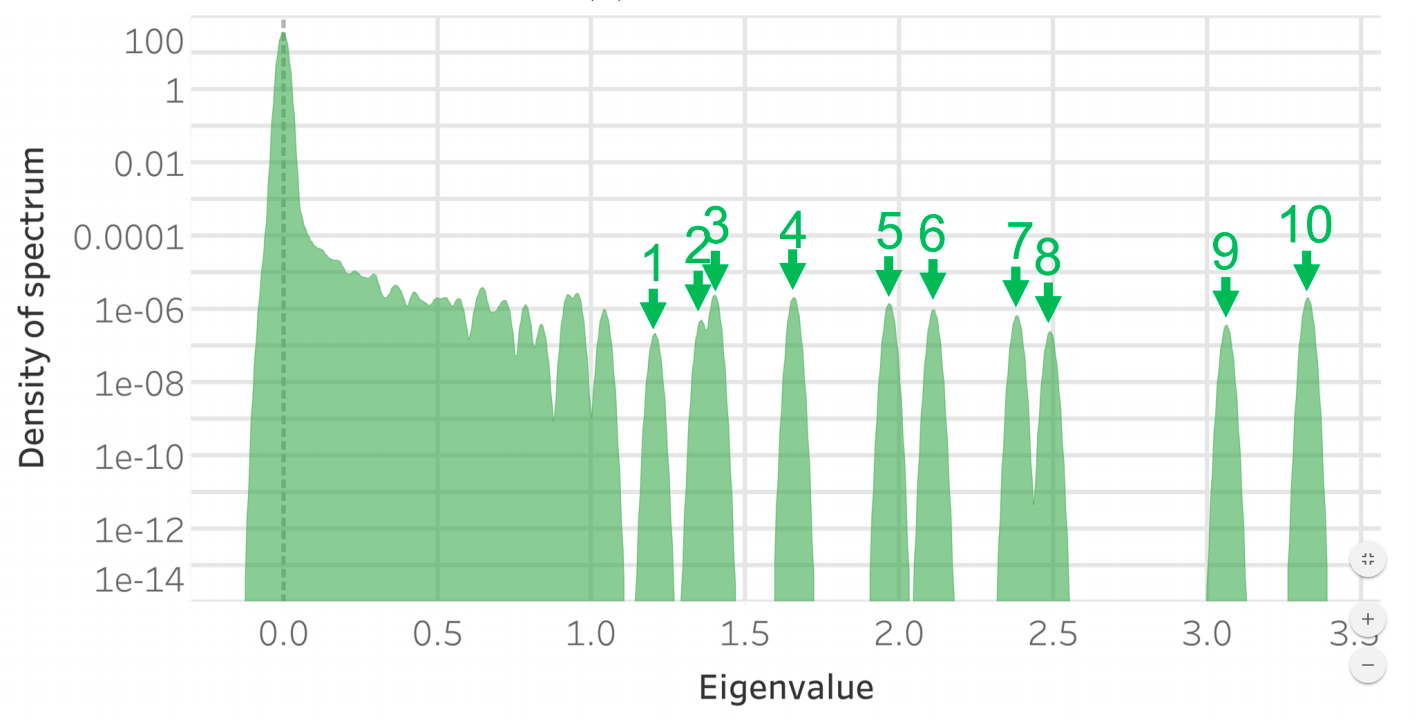}}
        \qquad
        \subfloat[Spectrum of test Hessian on MNIST\label{fig:ResNet-110_no}]{\includegraphics[width=0.45\linewidth]{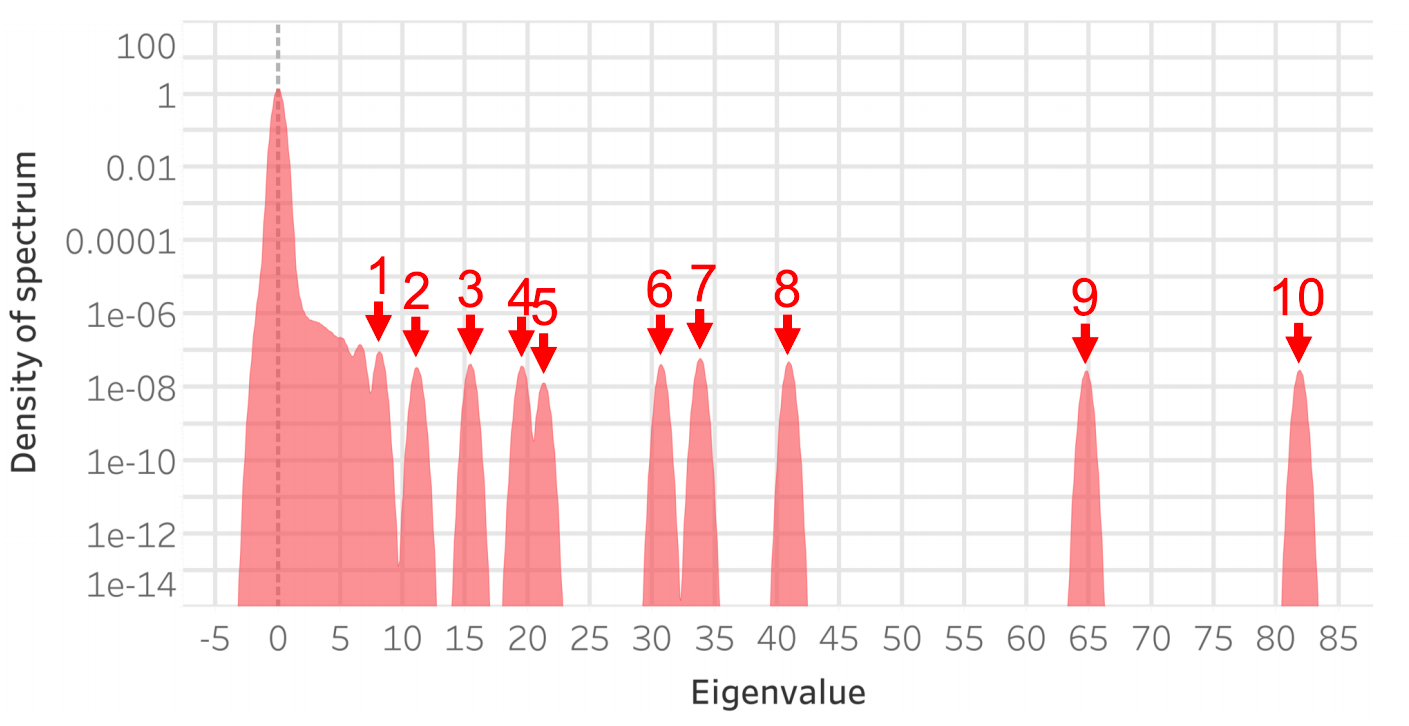}}
        \qquad
        \subfloat[Spectrum of test Hessian on CIFAR-10\label{fig:DenseNet}]{\includegraphics[width=0.45\linewidth]{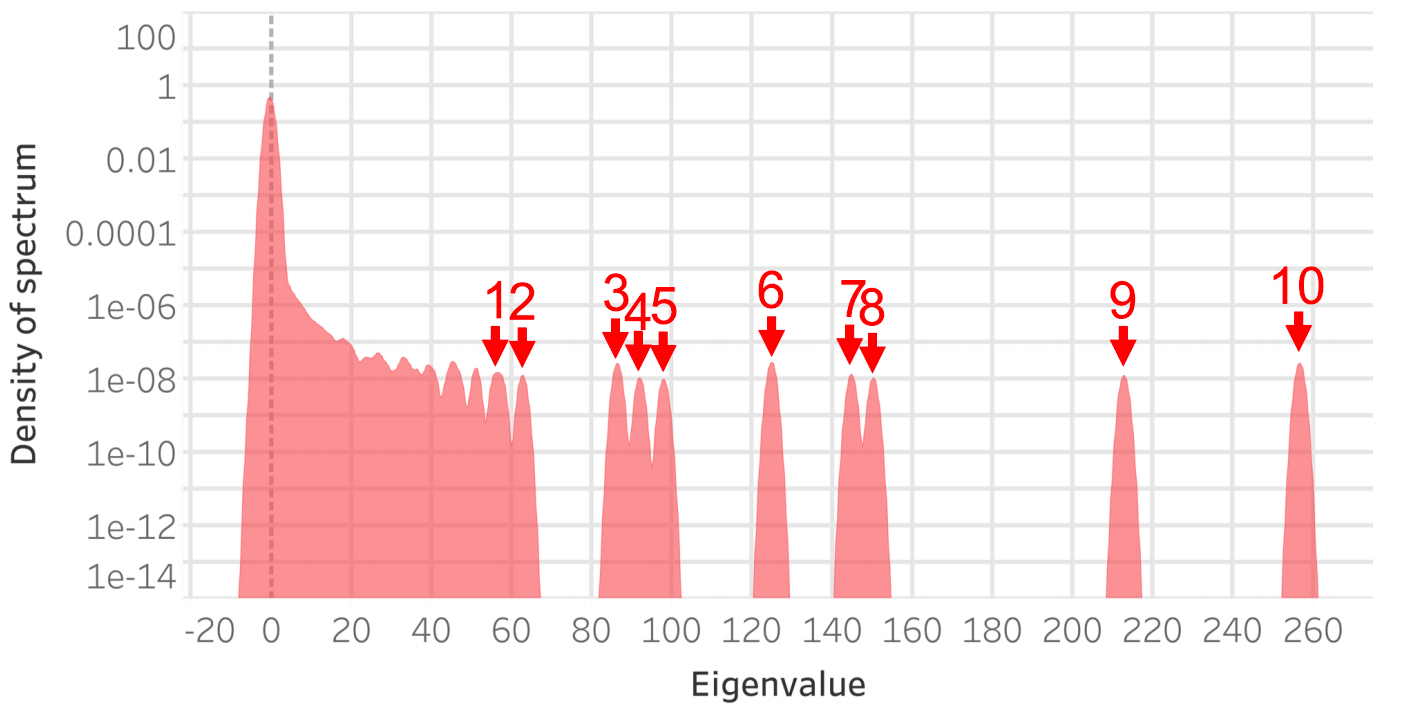}}
    \caption{The spectrums of train and test Hessian matrices on MNIST and CIFAR10. See \cite{papyan2018full}.}
    \label{fig:spectrum_Hessian}
\end{figure}

The property (2) by He {\it et al.} \cite{he2020piecewise} explains an important property of {\it mode connectivity}: the solutions found by stochastic gradient descent or its variants are connected with each other by a path in the weight space, on which all points have almost the same empirical risk. This property is found in two empirical works by Garipov {\it et al.} \cite{garipov2018loss} and Draxler {\it et al.} \cite{pmlr-v80-draxler18a}. A recent theoretical work by Kuditipudi {\it et al.} \cite{kuditipudi2019explaining} proves that  the mode connectivity can be guaranteed by dropout stability and noise stability. 

\textbf{Correspondence between the landscapes of empirical risk and expected risk.} Two works by Zhou and Feng \cite{zhou2017landscape} and Mei {\it et al.} and \cite{mei2018landscape} prove a linkage between the landscapes of the empirical risk surface and expected risk. This correspondence highlights the possibility of employing the results on the geometry of loss surfaces (empirical risk surfaces) to understand of generalizability (the gap between expected risks and empirical risks). Specifically, the authors prove that all of (1) the gradient of empirical risk, (2) the stationary points of the empirical risk, and (3) the empirical risk itself can be asymptotically approximated by their population counterparts. Moreover, they prove a generalization bound for non-convex case: at least by probability $1 - \delta$,
\begin{equation*}
\mathcal O \left ( \tau \sqrt{\frac{9}{8} [1 + c_r (D - 1)]} \sqrt{\frac{s \log (mU/D) + \log (4/\delta)}{m}} \right),
\end{equation*}
where all datums $x$ are assumed to be $\tau$-sub-Gaussian, $c_r = \max \left( r^2/16, \left(r^2/16\right)^{l-1} \right)$, and $s$ is the number of non-zero components in the weights. Later, \cite{mei2018landscape} prove a similar correspondence between the Hessian matrices of the expected risk and the empirical risk under an assumption that the sample size is larger than the parameter size.



\textbf{Eigenvalues of the Hessian.} Sagun {\it et al.} in 2016 \cite{sagun2016singularity} and 2018 \cite{sagun2018empirical} and Papyan in 2018 \cite{papyan2018full} conduct experiments to study the the eigenvalues of the Hessian of the loss surface. Sagun {\it et al.} \cite{sagun2016singularity, sagun2018empirical} discover that: (1) a large bulk of eigenvalues are centered close to zero; and (2) several outliers are located away from the bulk. Papyan \cite{papyan2018full} gives the full spectrums of the Hessian matrix. Fig. \ref{fig:spectrum_Hessian} compares among the spectrum of Hessian matrices when training and testing a VGG-11 on the datasets MNIST and CIFAR-10 (originally from a recent empirical work by \cite[p. 2, figs. 1(a) and 1(c) and p. 3, figs. 2(a) and 2(c)]{papyan2018full}).

\section{Reflecting the role of over-parameterization: Is it only harmful?}
\label{sec:over-parameterization}

One might blame the over-parameterization of deep learning for its lack of theoretical foundations. However, recent works also discover that such over-parameterization would have also contributed to the success of deep learning.

\textbf{Neural tangent kernel.} Neal in 1995 \cite{neal1995bayesian} and 1996 \cite{neal1996priors}, Williams \cite{williams1996computing}, and Hazan and Jaakkola\cite{hazan2015steps} prove that an infinite-width shallow neural network is equivalent to a Gaussian process. Damianou and Lawrence \cite{damianou2013deep}, Duvenaud {\it et al.} \cite{duvenaud2014avoiding}, Hensman and Lawrence \cite{hensman2014nested}, Lawrence and Moore \cite{lawrence2007hierarchical}, Bui {\it et al.} \cite{bui2016deep}, and Lee {\it et al.} \cite{lee2018deep}, gradually extend the equivalence to any infinite-width neural network of an arbitrary depth. Lee {\it et al.} \cite{lee2019wide} further prove that infinite-width neural networks of any depth behaves as linear models during training with gradient descent. Jacot {\it et al.} \cite{jacot2018neural} prove that the learned model for the least-squares regression problem during training is characterized by a linear differential equation the setting under the infinite-width regime.

\textbf{Influence of over-parameterization on the loss surface.} The over-parameterization would significantly reshape the loss surfaces of neural networks. Choromanska {\it et al.} \cite{choromanska2015loss} conducted an experiment that shows: (1) a large proportion of the local minima on the loss surface of an over-parameterized network are ``equivalent'' with each other (they have the same empirical risk); and (2) neural networks with small or normal sizes have spurious local minima but the proportion of the spurious local minima decreases rapidly when the network size increase. Li {\it et al.} \cite{li2018over} prove that over-parameterized fully-connected deep neural networks do not have any spurious local minimum under assumptions: (1) the activation functions are continuous; and (2) the loss functions are convex and differentiable. Nguyen {\it et al.} \cite{nguyen2018on} extend the ``no bad local minima'' property to networks trained under the cross-entropy loss. Nguyen \cite{nguyen2019on} further discover that the global minima are connected with each other and concentrated on a unique valley if the neural networks are sufficiently over-parameterized. 

\textbf{Influence of over-parameterization on the optimization.} It has been shown that the over-parameterization contribute to securing the optimization performance of gradient-based optimizers. Du {\it et al.} \cite{du2018gradient1} prove that gradient descent converges to global local minima under the quadratic loss at a linear convergence rate for two-layer neural networks as long as the network widths are sufficiently large. The authors also suggest that the over-parameterization can restrict the weights in all iterations close to the random initialization. Chizat and Bach \cite{chizat2018global} model the optimization of one-hidden-layer networks as minimizing a convex function of a measure  discretized into a mixture of particles by continuous-time gradient descent. They prove that the gradient flow characterizing the optimization converges to the global minimizers.

Going beyond one hidden layer, Allen-Zhu {\it et al.} \cite{pmlr-v97-allen-zhu19a}, Du {\it et al.} \cite{du2019gradient}, and Zou {\it et al.} \cite{zou2020gradient} concurrently prove that SGD converge to global optimal solution of over-parameterized deep neural networks in polynomial time with under slightly different assumptions on the network architectures and training data. Chen {\it et al.} \cite{chen2019much} prove that the convergence of optimization is guaranteed if the network width is polylogarithmic with the sample size $n$ and $1 / \epsilon$ where $\epsilon$ is the targeted error.

\textbf{Influence of over-parameterization on the generalization and learnability.} Andoni {\it et al.} \cite{andoni2014learning} study two-layer neural networks under assumptions that the node in the first layer are all linear functions while the hidden nodes in the last layer are non-linear. The authors prove that sufficiently wide neural network can learn any low-degree polynomial functions when trained by generic gradient descent algorithm and all weights are randomly initialized. Brutzkus {\it et al.} \cite{brutzkus2018sgd} also give generalization guarantees for two-layer neural networks under some assumptions. Li and Yuan \cite{li2017convergence} prove that over-parameterized one-hidden-layer ReLU networks have guaranteed small generalization error if they are trained by SGD for multi-class classification problems from random initialization. Also, the data is assumpted to be a mixture of well-separated distributions.

Arora {\it et al.} \cite{arora2019fine} and Allen-Zhu {\it et al.} \cite{allen2019learning} prove that the sample complexity for neural networks with two or three layers is almost independent with the parameter size. Chen {\it et al.} \cite{chen2019much} also show that the generalization is guaranteed if the network width is polylogarithmic with the sample size $n$ and $\epsilon^{-1}$ where $\epsilon$ is the targeted error. Cao {\it et al.} \cite{cao2019generalization} also prove generalization bounds for wide and deep neural networks. Wei {\it et al.} \cite{wei2019regularization} prove that regularizers can significantly reshape the generalization and optimization. 

\textbf{Influence of the network depth on the generalizability.} Canziani {\it et al.} \cite{canziani2016analysis} suggest that deeper neural networks could have better generalizabilities, summarizing the applications of neural networks in practice. This plays as a major part in the irreconciliation between the over-parameterization of neural networks and their excellent generalizabilities. Besides the previous results, another possible explanation is from information theory. Zhang {\it et al.} \cite{zhang2018information} prove a generalization bound that characterizes how the generalizability evolves when the network becomes deeper, based on some techniques developed by Xu and Raginsky \cite{xu2017information}.

\section{Theoretical foundations for specific architectures}
\label{sec:special_structures}

In the previous sections, we discuss the generalizabilities of neural networks in general. This section discusses the influence of some specific network structures.

\textbf{Convolutional neural networks.} Convolutional neural networks (CNNs) introduce convolutional layers into deep learning. It has been widely employed in computer vision \cite{krizhevsky2012imagenet}, natural language process \cite{yu2018qanet}, and deep reinforcement learning \cite{silver2016mastering}. 

Du {\it et al.} \cite{du2018many} study several statistical properties of CNNs. The authors prove that for an $c$-dimensional convolutional filter with linear activation needs $\tilde{\mathcal O} \left( c/\epsilon^2 \right)$ training examples for achieving a population prediction error $\epsilon$ on $d_X$-dimensional input, while a fully-connected counterpart needs $\Omega \left( d/\epsilon^2 \right)$ training examples. Moreover, the authors prove that a one-hidden-layer CNN needs $\tilde{\mathcal O} \left( (c + r)/\epsilon^2 \right)$ training examples when the output dimension is $r$ and the ratio of stride size to filter size is fixed. Zhou and Feng \cite{zhou2018understanding} prove (1) an $\mathcal O \left( \log \left(\prod_{i = 0}^D b_i \right) \right)$ generalization bound for CNN, where $b_i$ is the magnitude of the parameter in the $i$-th layer; and (2) one-to-one correspondence between convergence guarantees between the stationary points on the empirical risk landscape and their population counterparts. Lin and Zhang \cite{lin2019generalization} studies how the sparsity and permutation of convolutional layers would influence the spectral norm and then the generalizability.

Several works have also studied the generalizability of CNNs with residual connections \cite{he2016deep}, including He {\it et al.} \cite{he2020resnet} and Chen {\it et al.} \cite{chen2019theoretical}.

\textbf{Recurrent neural networks.} Recurrent Neural Networks (RNNs) possess a recurrent structure and have demonstrated promising performance in processing sequential data analysis, and have been deployed in many real-world areas, including natural language processing \cite{bahdanau2015neural, sutskever2014sequence} and speech recognition \cite{graves2006connectionist, graves2013speech}.  Chen {\it et al.} \cite{chen2019generalization} develops a generalization bound for RNNs based on works by Neyshabur {\it et al.} \cite{neyshabur2017pac} and Bartlett {\it et al.} \cite{bartlett2017spectrally}. Allen-Zhu and Li \cite{allen2019can} also analyze the generalizability of RNNs.

Tu {\it et al.} \cite{tu2020understanding} develop a gradient measure based on Fisher-Rao norm, which delivers a generalization bound as follows.
\begin{theorem}
Fix margin parameter $\alpha$, then for any $\delta> 0$, with probability at least $1-\delta$, the following holds for every RNN whose weight matrices $\theta=(U, W, V)$ satisfy $||V^T||_1\leq \beta_{V}, ||W^T||_1\leq \beta_{W}, ||U^T||_1\leq\beta_{U}$ and $||\theta||_{fs}\leq r$:
\begin{align}
\label{eq88}
& \mathbb{E}[\bm{1}_{\mathcal{M}_{y_L}(x,y)\leq0}] \nonumber\\
\leq& \dfrac{4k}{\alpha}\left(\dfrac{r||X||_F}{2n}\sqrt{\dfrac{1}{\lambda_{min}(\mathbb{E}(xx^T))}}+\dfrac{1}{n}\beta_V\beta_U||X^T||_1\Lambda\right)  \nonumber\\
& + \dfrac{1}{n}\sum\bm{1}_{\mathcal{M}_{y_L}(x_i,y_i)\leq a} + 3\sqrt{\dfrac{log \frac{2}{\delta}}{2n}}.
\end{align}
\end{theorem}
The gradient measure considerably tightens the bound and characterizes the relationship between generalizability and trainability. It suggests adding noise to the input data to boost the generalizability. It also justifies the technique gradient clipping \cite{merity2018regularizing, gehring2017convolutional, peters2018deep}.

\textbf{Networks for permutation invariant/equivariant functions.} Zaheer {\it et al.} \cite{zaheer2017deep}, Cohen and Welling \cite{cohen2016group}, and Cohen {\it et al.} \cite{cohen2019general} adopt machine learning to tasks based on ``sets", which are usually permutation invariant or equivariant. These models are then applied to a variety of scenarios, including 3D point cloud processing \cite{li2018pointcnn}, chemistry \cite{faber2016machine}, and astronomy \cite{ntampaka2016dynamical, ravanbakhsh2016estimating}. Sannai {\it et al.} \cite{sannai2019universal} prove a universal approximation theorem for this family of networks. Moreover, Sannai {\it et al.} \cite{sannai2019universal} show that the {\it free} parameters in this network family are exponentially fewer than usual models, which further suggests a significantly better generalizability. Sannai and Imaizumi \cite{sannai2019improved} prove that the deep model in Zaheer. {\it et al.} \cite{zaheer2017deep} can improve generalization bound by $\sqrt{n!}$, where $n$ is the number of  permuting coordinates of data. Brutzkus and Globerson \cite{brutzkus2019larger} also demonstrate that larger model size can boost the generalizability from both theoretical and empirical aspects.

\section{Looking beyond: Rising concerns of ethics and security}
\label{sec:ethics_security}

Rising concerns are seen on the ethical and security issues of deep learning, including privacy preservation, fairness protection, and adversarial robustness. Understanding the theoretical foundations of such issues are also of high importance. This section summarizes the recent advances in these areas.

\subsection{Privacy preservation}

Deep learning has been deployed to process massive personal data, including financial and medical records. It is of profound importance to discover the high-value population knowledge while protecting the highly sensitive privacy \cite{dwork2013s, dwork2014algorithmic}. 

\textbf{Differential privacy.} A major measure in measuring the privacy-preserving ability of an algorithm is differential privacy \cite{dwork2014algorithmic}. We call a learning algorithm is $(\varepsilon, \delta)$-differential private if when the training data is exposed to some small disturbance, the change in the output hypothesis is restricted as follows,
\begin{equation}  
	\label{eq:dp}
	\log \left[ \frac{\mathbb P_{\mathcal{A}(S)}(\mathcal{A}(S)\in B) - \delta}{\mathbb P_{\mathcal{A}(S')}(\mathcal{A}(S')\in B)} \right] \le \varepsilon,
\end{equation}
where $B$ is an arbitrary subset of the hypothesis space and $(S, S')$ is a neighboring sample set pair in which $S$ and $S'$ only differ by one example. The left-hand side of eq. (\ref{eq:dp}) is also called {\it privacy loss}.

In eq. (\ref{eq:dp}), we use division operation to measure the change of the output hypothesis. Through modifying the division operation or inserting some assumptions, many variants of notions for measuring privacy-preserving ability are proposed: (1) {\it concentration differential privacy} assumes that the privacy lossis sub-Gaussian \cite{dwork2016concentrated, bun2016concentrated}; (2) {\it mutual-information differential privacy} and {\it KL differential privacy} replace the division operation in privacy loss by mutual information and KL divergence, respectively \cite{cuff2016differential, wang2016relation, liao2017hypothesis, chaudhuri2019capacity}; (3) {\it R\'enyi differential privacy} further replaces the KL divergence by R\'enyi divergence \cite{mironov2017renyi, geumlek2017renyi}; etc.

These techniques have been also employed for privacy preservation in deep learning \cite{abadi2016deep}.

\textbf{Generalization-privacy relationship.} Some works have suggested that the privacy-preserving ability measured by differential privacy is almost equivalent to the generalizability.

Dwork {\it et al.} \cite{dwork2015preserving} prove a high-probability generalization bound for $(\varepsilon, \delta)$-differentially private machine learning algorithms as follows,
\begin{equation*}
\mathbb P\left[\mathcal R(\mathcal A(S)) - \hat{\mathcal R}_S (\mathcal A(S)) < 4\varepsilon\right] > 1- 8\delta^\varepsilon.
\end{equation*}

Oneto {\it et al.} \cite{oneto2017differential} improves the generalization bound as below
\begin{align*}
& \mathbb P\left[ \text{Diff } \mathcal R < \sqrt{6 \hat{ \mathcal{R}}_S(\mathcal{A}(S)) }\hat \varepsilon+6\left(\varepsilon^{2}+1 / N\right)\right] > 1 - 3 e^{-N \varepsilon^{2}},
\end{align*}
and
\begin{align*}
& \mathbb P\left[\text{Diff } \mathcal R < \sqrt{4 \hat{V}_S(\mathcal{A}(S))} \hat \varepsilon + \frac{5 N}{N-1}\left(\varepsilon^{2}+1 / N\right)\right] > 1 - 3 e^{-N \varepsilon^{2}},
\end{align*}
where
\begin{gather*}
\text{Diff } \mathcal R = \mathcal R(\mathcal A(S)) - \hat{\mathcal R}_S (\mathcal A(S)),\\
\hat \varepsilon = \varepsilon+\sqrt{1 / N},
\end{gather*}
and $\hat{V}_S(\mathcal{A}(S))$ is the empirical variance of $l(\mathcal{A}(S), \cdot)$:
\begin{gather*}
\hat{V}_S(\mathcal{A}(S)) = \frac{1}{2N(N-1)}\sum_{i\ne j} \left[\ell\left(\mathcal{A}(S), z_{i}\right)-\ell\left(\mathcal{A}(S), z_{j}\right)\right]^{2}.
\end{gather*}

Nissim and Stemmer \cite{nissim2015generalization} further prove that
\begin{equation*}
\mathbb P\left[\mathcal R(\mathcal A(S)) - \hat{\mathcal R}_S (\mathcal A(S)) <13\varepsilon\right] > 1 - \frac{2 \delta}{\varepsilon} \log \left(\frac{2}{\varepsilon}\right).
\end{equation*}

The tightest generalization bound by far is given by He {\it et al.} \cite{he2020tighter} as follows,
\begin{align}
\label{eq:high_probability_privacy}
& \mathbb{P}\left[\left|\hat{\mathcal{R}}_S(\mathcal{A}(S)) - \mathcal{R}(\mathcal{A}(S))\right| < 9\varepsilon\right] 
>  1-\frac{e^{-\varepsilon}\delta}{\varepsilon} \ln \left(\frac{2}{\varepsilon}\right).
\end{align}
This generalization bound is delivered in three stages. The authors first prove an on-average generalization bound for any $(\varepsilon,\delta)$-differentially private multi-database learning algorithm 
	\begin{equation*}
	\tilde{\mathcal{A}}: \vec{S} \to \mathcal H\times \{1,\cdots,k\},
	\end{equation*}
as follows, 	
	\begin{align}
	\label{eq:stab_databases}
	& \left|\underset{\vec{S} \sim \mathcal{D}^{N}}{\mathbb{E}}\left[\underset{ \mathcal{A}(\vec{S})}{\mathbb{E}}\left[\hat{\mathcal{R}}_{S_{i_{\mathcal{A}(\vec{S})}}}\left(h_{\mathcal{A}(\vec{S})}\right)\right] - \underset{ \mathcal{A}(\vec{S})}{\mathbb{E}}\left[\mathcal{R}\left(h_{\mathcal{A}(\vec{S})}\right)\right]\right]\right| 
	 \leq e^{-\varepsilon}k \delta+1-e^{-\varepsilon},
	\end{align}
	where the loss function $\Vert l\Vert_{\infty}\le 1$.

They then obtain a high-probability generalization bound for multi-database algorithms as follows,
		\begin{equation}
		\label{eq:generalization_bound_conflict}
		\mathbb{P}\left[\hat{\mathcal{R}}_{S_{i_{\mathcal{A}(\vec{S})}}}\left(h_{\mathcal{A}(\vec{S})}\right) \leq \mathcal{R}\left(h_{\mathcal{A}(\vec{S})}\right)+k e^{-\varepsilon} \delta+ 3\varepsilon\right] \geq \varepsilon.
		\end{equation}
		
They eventually prove eq. (\ref{eq:high_probability_privacy}) by reduction to absurdity.


\subsection{Algorithmic fairness}

Deep learning been deployed in many critical decision-making areas, such as mortgage approval \cite{ramadorai2017predictably}, credit card assessment \cite{khandani2010consumer, chen2019fairness}, college admission \cite{acikkar2009support, chen2014training}, employee selection \cite{faliagka2012application}, and recidivism prediction \cite{brennan2009evaluating}.

However, long-standing systematical discrimination has been observed in these areas against people based on their backgrounds, including races, genders, nationalities, ages, and religious. Additionally, people with minority backgrounds are institutionally under-represented in historical data. Meanwhile, it has been revealed in an increasing number of reports that the discrimination exists in the data are inherited and even intensified in the modeled learned by machine learning-based algorithms. The fairness concerns are particularly severe due to the black-box nature. For example, facial recognition is reportedly to be ``accurate, if you are a white guy'' \cite{lohr2018facial}; a police system for recidivism prediction, Correctional Offender Management Profiling for Alternative Sanctions (COMPAS) \cite{brennan2009evaluating}, tends to label minority people with significantly higher possibility to be reoffenders. Mitigrating the fairness risk is timely and important.

Many approaches have been presented to address the the fairness issues. They can be roughly categorized into three parts: (1) {\it pre-processing methods}, which aims to ``rectify'' data" before processed by learning algorithms \cite{calders2009building, feldman2015certifying, kamiran2012data, dwork2012fairness, calmon2017optimized, hardt2016equality, zemel2013learning, louizos2015variational}; (2) {\it post-processing methods} redress the outputs of learning systems to remove the potential discrimination \cite{woodworth2017learning}; and (3) {\it in-processing methods} adapt constraints or regularizers in the training procedures to enforce fairness \cite{zafar2017fairness, agarwal2018reductions, kamishima2012fairness, menon2018cost, kearns2018preventing}.

\subsection{Adversarial robustness}

Szegedy {\it et al.} \cite{szegedy2014intriguing} discover that neural networks are vulnerable to adversarial examples which are slightly different from true datums but are likely to be misclassified by neural networks. Since then, many approaches have been proposed on attacking neural networks by adversarial examples or defending the adversarial attacks \cite{biggio2013evasion, papernot2017practical, papernot2016limitations, papernot2016towards, goodfellow2014explaining, nguyen2015deep, papernot2016limitations, gilmer2018motivating}.

\textbf{Generalizability in the presence of adversarial examples.} Schmidt {\it et al.} \cite{schmidt2018adversarially} first study the generalizability in the presence of adversarial examples. They define a new notion, {\it adversarial robustness generalization} for measuring the generalizability when adversarial examples exist in data. In contrast, the generalizability when there is no adversarial examples is evaluated by the {\it standard generalization}. They prove that to secure the same generalization, adversarial robustness generalization needs a higher sample size, when the data is drawn from some specific distributions and the adversarial examples are under the $l_\infty$ norm.

Cullina {\it et al.} \cite{cullina2018pac} extends the PAC-learning framework to learning with adversarial examples. The authors propose a {\it corrupted hypothesis classes} which is defined to be the hypothesis classes for the classification problem in the presence of adversaries. The VC dimension of a corrupted hypothesis class is then defined to be {\it adversarial VC-dimension}, while the {\it standard VC-dimension} is defined on clean datasets. They show that the adversarial VC-dimension can be either larger or smaller than the standard VC-dimension. Montasser {\it et al.} \cite{montasser2019vc} further define an {\it agnostic robust PAC learnability} and {\it realizable robust PAC learnability}. When a learning algorithm whose rule meets the requirements of both agnostic robust PAC learnability and realizable robust PAC learnability, the learning is defined to be improper. They then show that under an improper learning rule, any hypothesis space $\mathcal H$ is robustly PAC learnable, if its VC dimension is finite.

\textbf{Generalizability of adversarial training.} Amongst the defense approaches, adversarial training \cite{dai2018adversarial, li2018second, baluja2018learning, zheng2019distributionally} can considerably improve the adversarial robustness of deep neural networks against adversarial examples. Specifically, adversarial training can be formulated as solving the following minimax-loss problem \cite{tu2019theoretical, yin2019rademacher, khim2018adversarial},
\begin{equation*}
    \min_{\theta} \frac{1}{N} \sum_{i=1}^N
    \max_{\Vert x_{i}^\prime - x_{i} \Vert \leq \rho}
    l (h_{\theta} (x_i^\prime), y_i),
\end{equation*}
where $h_{\theta}$ is the hypothesis parameterized by $\theta$, $N$ is the training sample size, $x_i$ is a feature, $y_i$ is the corresponding label, and $l$ is the loss function. Intuitively, adversarial training optimizes neural networks according to the performance on worst-case examples, which are most likely to be adversarial examples.

Based on the minimax loss framework, several generalization bounds have been proposed \cite{yin2019rademacher, tu2019theoretical, khim2018adversarial}. All these results suggest that adversarial training would comprise generalizability when improving the adversarial robustness.

Yin {\it et al.} \cite{yin2019rademacher} studies how adversarial training would influence the generalizability based on the techniques developed by Bartlett {\it et al.} \cite{bartlett2017spectrally} and Raghunathan {\it et al.} \cite{raghunathan2018certified}. They prove a tight upper bound on the Rademacher complexity which leads to a generalization bound. However, the generalization bound is only for neural networks with one hidden layer and ReLU activation functions.

To overcome this shortcoming, Khim {\it et al.} \cite{khim2018adversarial} next prove a generalization bound via function transformation. They prove an upper bound for the Rademacher complexity via techniques developed by Golowich {\it et al.} \cite{golowich2017size}.

Tu {\it et al.} \cite{tu2019theoretical} develop another path for evaluating the generalizability: (1) they first show that one can remove the maximization operation in the objective function in the minimax framework, $\max_{\Vert x_{i}^\prime - x_{i} \Vert \leq \rho} l (h_{\theta} (x_i^\prime), y_i)$ via a re-parameterization mapping, while the parameter distribution would not shift much under the Wasserstein distribution; (2) they prove a local worst-case risk bound for the re-parameterized optimization problem which has been a minimization problem; and (3) they prove a generalization bound based on the local worst-case risk bound. This generalization bound is more generalization in two aspects: (1) the adversarial examples encompass all bounded adversaries under the $l_q$ norm where $l_q$ is arbitrary; (2) the approach can be applied to multi-class classification and almost all practical loss functions, including hinge loss and ramp loss.

On the other side, Bhagoji {\it et al.} \cite{bhagoji2019lower} also prove a lower bound for the generalization error of classifiers in the presence of adversarial examples. Min {\it et al.} \cite{min2020curious} discover that the performance of adversarial training does not always have a clear correlation with the training sample size. The authors categorize all learning problems in three parts: (1) in the {\it weak adversary regime}, an increasing training sample size corresponds to an improving generalizability; (2) in the {\it medium adversary regime}, one may observe a double descent curve of the generalization error, when the training sample size increases; and (3) in the {\it strong adversary regime}, more  training data causes the generalization error to increase. Here, the three regimes can be determined by the level of adversaries. Similar phenomena to the medium adversary regime have also been discovered by Chen {\it et al.} \cite{chen2020more}.

Other theoretical analyses include an improved generalization bound via Rademacher complexity by Attias {\it et al.} \cite{attias2019improved}, the trade-off between accuracy and robustness \cite{zhang2019theoretically}, a generalization bound via optimal transport and optimal couplings by Pydi and Jog \cite{pydi2020adversarial}.

\begin{table}[h]
  \caption{Comparisons of major generalization bounds for deep learning.}
  \label{table:generalization_bounds}
  \centering
  \footnotesize
  \begin{tabular}{ccccccccc}
    \toprule
					& Tool			& Generalization bound \\
    \midrule
	Goldberg and Jerrum \cite{goldberg1995bounding}	& VC-dimension	& $\mathcal O \left( \sqrt{\frac{\log (m W^2)}{m W^2}} \right)$\\
    \midrule
	Bartlett {\it et al.} \cite{bartlett1999almost}	& VC-dimension	& $\mathcal O \left( \sqrt{\frac{\log (m W L \log W +W L^2)}{m W L \log W +W L^2}} \right)$\\
    \midrule
	Harvey {\it et al.} \cite{harvey2017nearly}	& VC-dimension	& $\mathcal O \left( \sqrt{\frac{\log (m W U \log((d + 1)p))}{m W U \log((d + 1)p)}} \right)$\\
    \midrule
	Neyshabur {\it et al.} \cite{neyshabur2015norm}	& Rademacher complexity	& $\mathcal O\left( \frac{B2^D\prod_{j = 1}^DM_F(j)}{\sqrt m} \right)$\\
    \midrule
	Bartlett {\it et al.} \cite{bartlett2017spectrally}	& \tabincell{c}{Rademacher complexity \\ \& covering number}	& $\tilde{\mathcal O} \left( \frac{\|X\|_2  \log W}{\gamma m} \left( \prod_{i = 1}^D \rho_i \| A \|_\sigma  \right) \left( \sum_{i = 1}^D \frac{\|A_i^\top - M_i^\top\|_{2, 1}^{2/3}}{ \| A \|_\sigma^{2/3}}  \right)^{3/2} \right)$\\
    \midrule
	Neyshabur {\it et al.} \cite{neyshabur2017pac}	& PAC-Bayes 	& $\mathcal O \left( \sqrt{\frac{B^2 D^2 W \log (DW) \prod_{i = 1}^D \frac{\|W_i\|_F^2}{\|W_i\|_2^2} + \log \frac{D m}{\delta}}{\gamma^2 m}} \right)$\\
    \midrule
	Golowich {\it et al.} \cite{golowich2017size}	& Rademacher complexity 	& $\mathcal O\left( \frac{B \sqrt D\prod_{j = 1}^DM_F(j)}{\sqrt m} \right) \text{ or } \mathcal O\left( \frac{B \sqrt{D+\log m} \prod_{j = 1}^DM_F(j)}{\sqrt m} \right)$\\
    \midrule
	Dieuleveut {\it et al.} \cite{dieuleveut2016nonparametric}	& SGM on convex loss 	& $\mathcal O\left( m^{-\frac{2 \min \{\zeta, 1\}}{2 \min \{\zeta, 1\} + \gamma}} \right)$\\
    \midrule
	Hardt {\it et al.} \cite{hardt2016train}	& \tabincell{c}{SGM on non-convex loss \\ \& algorithmic stability}	& $\frac{1 + 1/\beta c}{m - 1} (2cL^2)^{\frac{1}{\beta c + 1}} T^{\frac{1}{\beta c + 1}}$\\
    \midrule
	Raginsky {\it et al.} \cite{raginsky2017non}	& \tabincell{c}{SGM on non-convex loss \\ \& algorithmic stability} 	& $\tilde{\mathcal O} \left( \frac{\beta ( \beta + U )^2}{\lambda_*} \delta^{1/4} \log \left( \frac{1}{\epsilon} + \epsilon \right) + \frac{(\beta + U)^2}{\lambda_* + m} + \frac{U \log(\beta + 1)}{\beta} \right)$\\
    \midrule
	Pensia {\it et al.} \cite{pensia2018generalization}	& \tabincell{c}{SGM on non-convex loss \\ \& information theory} 	& $\tilde{\mathcal O} \left( \sqrt{\frac{R^2}{m} \sum_{t = 1}^T \frac{\eta_t^2 L^2}{\sigma^2_t}} \right)$\\
    \midrule
	Mou {\it et al.} \cite{mou2017generalization}	& \tabincell{c}{SGM on non-convex loss \\ \& algorithmic stability} 	& $\frac{2LC}{m} \left( \beta \sum_{t = 1}^N \eta_t \right)^2$\\
    \midrule
	Mou {\it et al.} \cite{mou2017generalization}	& \tabincell{c}{SGM on non-convex loss \\ \& PAC-Bayes} 	& $\mathcal O \left( \sqrt{\frac{\beta}{m} \sum_{k = 1}^N \eta_k e^{-\frac{\lambda}{3}(T_N - T_k)} \mathbb E[\| g_k(\theta_k) \|^2]} \right)$\\
    \midrule
	Lever {\it et al.} \cite{lever2013tighter}	& \tabincell{c}{Distribution-dependent prior \\ \& PAC-Bayes} 	& $\frac{C^*}{Cm} \left( \lambda \sqrt{\frac{2}{m} \log\frac{2\xi(m)}{\delta}} + \frac{\gamma^2}{2m} + \log \frac{2}{\delta} \right)$\\
    \midrule
	Li. {\it et al.} \cite{li2019generalization}	& \tabincell{c}{Distribution-dependent prior \\ \& Bayes-stability} 	& $\mathcal O \left( \frac{C}{m} \sqrt{\mathbb E_S \left[ \sum_{t=1}^T \frac{\eta^2_t}{\sigma_t^2} g_e(t) \right]} \right)$\\
    \midrule
	Negrea {\it et al.} \cite{negrea2019information}	& \tabincell{c}{Data-dependent prior \\ \& information theory} 	& $\sqrt{\frac{2 \sigma^2}{m - n} I^{S_J, U}(\theta; S_J^c)} \le \sqrt{\frac{2 \sigma^2}{m - n} \mathbb E^{S_J, U} [\text{KL}(Q||P)]}$\\
    \bottomrule
  \end{tabular}
\end{table}

\section{Summary and discussion}

In this paper, we review the literature on the deep learning theory from six aspects: (1) complexity and capacity-based method for analyzing the generalizability of deep learning; (2) stochastic differential equations and their dynamic systems for modelling stochastic gradient methods (SGM), which characterize the implicit regularization introduced by the limited exploration space; (3) the geometry of the loss landscape governing the trajectories of the dynamic systems; (4) the roles of over-parameterization of deep neural networks from both positive and negative perspectives; (5) parallel to the previous parts, the theoretical foundations for some special structures in the neural network architecture; 
and (6) the rising concerns on ethics and security and their relationships with deep learning theory.

Many works give upper bounds for the generalization error of deep learning. The results are summarized in Table \ref{table:generalization_bounds}.

From our view, these six directions are promising in establishing the theoretical foundations of deep learning. However, the current ``states of the art'' are still far away from sufficient; all of them are facing major difficulties. In this paper, we list three main directions for future works:
\begin{itemize}
\item

The current measures for the complexity evaluate the whole hypothesis space, but can hardly capture the complexity of the ``effective'' hypothesis space that the training algorithm would output. Thus, a ``local'' version of complexity would be helpful; see an example \cite{bartlett2005local}.

\item
The current models for SGM usually ignore some of the  popular training techniques, such as momentum, adaptive learning rate, gradient clipping, batch normalization, etc. Some works have been studying on thesis problems, but a full ``landscape'' is still missing. Also, the works are usually assuming the gradient noise is drawn from some specific distributions, such as Gaussian distributions or $\alpha$-stable distributions.

\item
The geometry of the loss surface remains premature. The extreme complexity of neural networks makes simple explanations unaccessible. Efforts on understanding some sophisticated geometrical properties are demanded, such as the sharpness/flatness of the local minima, the potential categorization of the local minima (valley) according to their performance, and the volumes of the local minima from different categories.

\end{itemize}

\section*{Acknowledgements}

The authors appreciate Weijie J. Su and Yu Yao for the helpful discussions.

\bibliographystyle{abbrv}
\bibliography{survey}

\begin{thebibliography}{100}

\bibitem{abadi2016deep}
M.~Abadi, A.~Chu, I.~Goodfellow, H.~B. McMahan, I.~Mironov, K.~Talwar, and
  L.~Zhang.
\newblock Deep learning with differential privacy.
\newblock In {\em ACM SIGSAC Conference on Computer and Communications
  Security}, pages 308--318, 2016.

\bibitem{acikkar2009support}
M.~Acikkar and M.~F. Akay.
\newblock Support vector machines for predicting the admission decision of a
  candidate to the school of physical education and sports at cukurova
  university.
\newblock {\em Expert Systems with Applications}, 36(3):7228--7233, 2009.

\bibitem{agarwal2018reductions}
A.~Agarwal, A.~Beygelzimer, M.~Dud{\'\i}k, J.~Langford, and H.~Wallach.
\newblock A reductions approach to fair classification.
\newblock {\em arXiv preprint arXiv:1803.02453}, 2018.

\bibitem{ahn2012bayesian}
S.~Ahn, A.~Korattikara, and M.~Welling.
\newblock {Bayesian} posterior sampling via stochastic gradient {Fisher}
  scoring.
\newblock {\em arXiv preprint arXiv:1206.6380}, 2012.

\bibitem{aizenberg2000multi}
I.~Aizenberg, N.~N. Aizenberg, and J.~P. Vandewalle.
\newblock {\em Multi-Valued and Universal Binary Neurons: Theory, Learning and
  Applications}.
\newblock Springer Science \& Business Media, 2000.

\bibitem{allen2019can}
Z.~Allen-Zhu and Y.~Li.
\newblock Can {SGD} learn recurrent neural networks with provable
  generalization?
\newblock In {\em Advances in Neural Information Processing Systems}, pages
  10331--10341, 2019.

\bibitem{allen2019learning}
Z.~Allen-Zhu, Y.~Li, and Y.~Liang.
\newblock Learning and generalization in overparameterized neural networks,
  going beyond two layers.
\newblock In {\em Advances in Neural Information Processing Systems}, pages
  6158--6169, 2019.

\bibitem{pmlr-v97-allen-zhu19a}
Z.~Allen-Zhu, Y.~Li, and Z.~Song.
\newblock A convergence theory for deep learning via over-parameterization.
\newblock In {\em International Conference on Machine learning}, 2019.

\bibitem{alom2019state}
M.~Z. Alom, T.~M. Taha, C.~Yakopcic, S.~Westberg, P.~Sidike, M.~S. Nasrin,
  M.~Hasan, B.~C. Van~Essen, A.~A. Awwal, and V.~K. Asari.
\newblock A state-of-the-art survey on deep learning theory and architectures.
\newblock {\em Electronics}, 8(3):292, 2019.

\bibitem{andoni2014learning}
A.~Andoni, R.~Panigrahy, G.~Valiant, and L.~Zhang.
\newblock Learning polynomials with neural networks.
\newblock In {\em International Conference on Machine learning}, pages
  1908--1916, 2014.

\bibitem{anthony2009neural}
M.~Anthony and P.~L. Bartlett.
\newblock {\em Neural Network Learning: Theoretical Foundations}.
\newblock Cambridge University Press, 2009.

\bibitem{arora2019fine}
S.~Arora, S.~S. Du, W.~Hu, Z.~Li, and R.~Wang.
\newblock Fine-grained analysis of optimization and generalization for
  overparameterized two-layer neural networks.
\newblock {\em arXiv preprint arXiv:1901.08584}, 2019.

\bibitem{arpit2017closer}
D.~Arpit, S.~Jastrzebski, N.~Ballas, D.~Krueger, E.~Bengio, M.~S. Kanwal,
  T.~Maharaj, A.~Fischer, A.~Courville, and Y.~Bengio.
\newblock A closer look at memorization in deep networks.
\newblock In {\em International Conference on Machine learning}, 2017.

\bibitem{attias2019improved}
I.~Attias, A.~Kontorovich, and Y.~Mansour.
\newblock Improved generalization bounds for robust learning.
\newblock In {\em Algorithmic Learning Theory}, pages 162--183, 2019.

\bibitem{bahdanau2015neural}
D.~Bahdanau, K.~Cho, and Y.~Bengio.
\newblock Neural machine translation by jointly learning to align and
  translate.
\newblock In {\em International Conference on Learning Representations}, 2015.

\bibitem{baluja2018learning}
S.~Baluja and I.~Fischer.
\newblock Learning to attack: Adversarial transformation networks.
\newblock In {\em AAAI Conference on Artificial Intelligence}, volume~1,
  page~3, 2018.

\bibitem{bartlett1999generalization}
P.~Bartlett and J.~Shawe-Taylor.
\newblock Generalization performance of support vector machines and other
  pattern classifiers.
\newblock {\em Advances in Kernel methods—Support Vector Learning}, pages
  43--54, 1999.

\bibitem{bartlett2005local}
P.~L. Bartlett, O.~Bousquet, and S.~Mendelson.
\newblock Local rademacher complexities.
\newblock {\em The Annals of Statistics}, 33(4):1497--1537, 2005.

\bibitem{bartlett2017spectrally}
P.~L. Bartlett, D.~J. Foster, and M.~J. Telgarsky.
\newblock Spectrally-normalized margin bounds for neural networks.
\newblock In {\em Advances in Neural Information Processing Systems}, pages
  6240--6249, 2017.

\bibitem{bartlett1999almost}
P.~L. Bartlett, V.~Maiorov, and R.~Meir.
\newblock Almost linear {VC} dimension bounds for piecewise polynomial
  networks.
\newblock In {\em Advances in Neural Information Processing Systems}, pages
  190--196, 1999.

\bibitem{bartlett2002rademacher}
P.~L. Bartlett and S.~Mendelson.
\newblock Rademacher and {Gaussian} complexities: Risk bounds and structural
  results.
\newblock {\em Journal of Machine Learning Research}, 3(Nov):463--482, 2002.

\bibitem{bartlett1996vc}
P.~L. Bartlett and R.~C. Williamson.
\newblock The {VC} dimension and pseudodimension of two-layer neural networks
  with discrete inputs.
\newblock {\em Neural Computation}, 8(3):625--628, 1996.

\bibitem{baum1989size}
E.~B. Baum and D.~Haussler.
\newblock What size net gives valid generalization?
\newblock In {\em Advances in Neural Information Processing Systems}, pages
  81--90, 1989.

\bibitem{bengio2006greedy}
Y.~Bengio, P.~Lamblin, D.~Popovici, and H.~Larochelle.
\newblock Greedy layer-wise training of deep networks.
\newblock {\em Advances in Neural Information Processing Systems}, 19:153--160,
  2006.

\bibitem{bhagoji2019lower}
A.~N. Bhagoji, D.~Cullina, and P.~Mittal.
\newblock Lower bounds on adversarial robustness from optimal transport.
\newblock In {\em Advances in Neural Information Processing Systems}, pages
  7498--7510, 2019.

\bibitem{biggio2013evasion}
B.~Biggio, I.~Corona, D.~Maiorca, B.~Nelson, N.~{\v{S}}rndi{\'c}, P.~Laskov,
  G.~Giacinto, and F.~Roli.
\newblock Evasion attacks against machine learning at test time.
\newblock In {\em European Conference on Machine Learning}, 2013.

\bibitem{blei2017variational}
D.~M. Blei, A.~Kucukelbir, and J.~D. McAuliffe.
\newblock Variational inference: A review for statisticians.
\newblock {\em Journal of the American statistical Association},
  112(518):859--877, 2017.

\bibitem{blumer1989learnability}
A.~Blumer, A.~Ehrenfeucht, D.~Haussler, and M.~K. Warmuth.
\newblock Learnability and the {Vapnik-Chervonenkis} dimension.
\newblock {\em Journal of the ACM}, 36(4):929--965, 1989.

\bibitem{bousquet2002stability}
O.~Bousquet and A.~Elisseeff.
\newblock Stability and generalization.
\newblock {\em Journal of Machine Learning Research}, 2(Mar):499--526, 2002.

\bibitem{brennan2009evaluating}
T.~Brennan, W.~Dieterich, and B.~Ehret.
\newblock Evaluating the predictive validity of the compas risk and needs
  assessment system.
\newblock {\em Criminal Justice and Behavior}, 36(1):21--40, 2009.

\bibitem{brutzkus2019larger}
A.~Brutzkus and A.~Globerson.
\newblock Why do larger models generalize better? a theoretical perspective via
  the xor problem.
\newblock In {\em International Conference on Machine learning}, pages
  822--830, 2019.

\bibitem{brutzkus2018sgd}
A.~Brutzkus, A.~Globerson, E.~Malach, and S.~Shalev-Shwartz.
\newblock {SGD} learns over-parameterized networks that provably generalize on
  linearly separable data.
\newblock In {\em International Conference on Learning Representations}, 2018.

\bibitem{bu2020tightening}
Y.~Bu, S.~Zou, and V.~V. Veeravalli.
\newblock Tightening mutual information based bounds on generalization error.
\newblock {\em IEEE Journal on Selected Areas in Information Theory}, 2020.

\bibitem{bui2016deep}
T.~Bui, D.~Hern{\'a}ndez-Lobato, J.~Hernandez-Lobato, Y.~Li, and R.~Turner.
\newblock Deep {Gaussian} processes for regression using approximate
  expectation propagation.
\newblock In {\em International Conference on Machine learning}, pages
  1472--1481, 2016.

\bibitem{bun2016concentrated}
M.~Bun and T.~Steinke.
\newblock Concentrated differential privacy: Simplifications, extensions, and
  lower bounds.
\newblock In {\em Theory of Cryptography Conference}, pages 635--658, 2016.

\bibitem{calders2009building}
T.~Calders, F.~Kamiran, and M.~Pechenizkiy.
\newblock Building classifiers with independency constraints.
\newblock In {\em IEEE International Conference on Data Mining Workshops},
  pages 13--18, 2009.

\bibitem{calmon2017optimized}
F.~Calmon, D.~Wei, B.~Vinzamuri, K.~Natesan~Ramamurthy, and K.~R. Varshney.
\newblock Optimized pre-processing for discrimination prevention.
\newblock {\em Advances in Neural Information Processing Systems},
  30:3992--4001, 2017.

\bibitem{canziani2016analysis}
A.~Canziani, A.~Paszke, and E.~Culurciello.
\newblock An analysis of deep neural network models for practical applications.
\newblock {\em arXiv preprint arXiv:1605.07678}, 2016.

\bibitem{cao2019generalization}
Y.~Cao and Q.~Gu.
\newblock Generalization bounds of stochastic gradient descent for wide and
  deep neural networks.
\newblock In {\em Advances in Neural Information Processing Systems}, pages
  10836--10846, 2019.

\bibitem{carlini2018secret}
N.~Carlini, C.~Liu, J.~Kos, {\'U}.~Erlingsson, and D.~Song.
\newblock The secret sharer: Measuring unintended neural network memorization
  \& extracting secrets.
\newblock {\em arXiv preprint arXiv:1802.08232}, 2018.

\bibitem{chaudhari2018stochastic}
P.~Chaudhari and S.~Soatto.
\newblock Stochastic gradient descent performs variational inference, converges
  to limit cycles for deep networks.
\newblock In {\em Information Theory and Applications Workshop}, 2018.

\bibitem{chaudhuri2019capacity}
K.~Chaudhuri, J.~Imola, and A.~Machanavajjhala.
\newblock Capacity bounded differential privacy.
\newblock {\em arXiv preprint arXiv:1907.02159}, 2019.

\bibitem{chen2018rise}
H.~Chen, O.~Engkvist, Y.~Wang, M.~Olivecrona, and T.~Blaschke.
\newblock The rise of deep learning in drug discovery.
\newblock {\em Drug Discovery Today}, 23(6):1241--1250, 2018.

\bibitem{chen2019theoretical}
H.~Chen, Z.~Mo, Z.~Yang, and X.~Wang.
\newblock Theoretical investigation of generalization bound for residual
  networks.
\newblock In {\em International Joint Conference on Artificial Intelligence},
  pages 2081--2087, 2019.

\bibitem{chen2019fairness}
J.~Chen, N.~Kallus, X.~Mao, G.~Svacha, and M.~Udell.
\newblock Fairness under unawareness: Assessing disparity when protected class
  is unobserved.
\newblock In {\em Conference on Fairness, Accountability, and Transparency},
  pages 339--348, 2019.

\bibitem{chen2014training}
J.-F. Chen and Q.~H. Do.
\newblock Training neural networks to predict student academic performance: A
  comparison of cuckoo search and gravitational search algorithms.
\newblock {\em International Journal of Computational Intelligence and
  Applications}, 13(01):1450005, 2014.

\bibitem{chen2020more}
L.~Chen, Y.~Min, M.~Zhang, and A.~Karbasi.
\newblock More data can expand the generalization gap between adversarially
  robust and standard models.
\newblock {\em arXiv preprint arXiv:2002.04725}, 2020.

\bibitem{chen2019generalization}
M.~Chen, X.~Li, and T.~Zhao.
\newblock On generalization bounds of a family of recurrent neural networks.
\newblock {\em arXiv preprint arXiv:1910.12947}, 2019.

\bibitem{chen2014stochastic}
T.~Chen, E.~Fox, and C.~Guestrin.
\newblock Stochastic gradient {Hamiltonian} {Monte Carlo}.
\newblock In {\em International Conference on Machine learning}, pages
  1683--1691, 2014.

\bibitem{chen2016statistical}
X.~Chen, J.~D. Lee, X.~T. Tong, and Y.~Zhang.
\newblock Statistical inference for model parameters in stochastic gradient
  descent.
\newblock {\em arXiv preprint arXiv:1610.08637}, 2016.

\bibitem{chen2018stability}
Y.~Chen, C.~Jin, and B.~Yu.
\newblock Stability and convergence trade-off of iterative optimization
  algorithms.
\newblock {\em arXiv preprint arXiv:1804.01619}, 2018.

\bibitem{chen2019much}
Z.~Chen, Y.~Cao, D.~Zou, and Q.~Gu.
\newblock How much over-parameterization is sufficient to learn deep {ReLU}
  networks?
\newblock {\em arXiv preprint arXiv:1911.12360}, 2019.

\bibitem{cherian2020efficient}
J.~J. Cherian, A.~G. Taube, R.~T. McGibbon, P.~Angelikopoulos, G.~Blanc,
  M.~Snarski, D.~D. Richman, J.~L. Klepeis, and D.~E. Shaw.
\newblock Efficient hyperparameter optimization by way of pac-bayes bound
  minimization.
\newblock {\em arXiv preprint arXiv:2008.06431}, 2020.

\bibitem{chizat2018global}
L.~Chizat and F.~Bach.
\newblock On the global convergence of gradient descent for over-parameterized
  models using optimal transport.
\newblock {\em Advances in Neural Information Processing Systems},
  31:3036--3046, 2018.

\bibitem{choromanska2015loss}
A.~Choromanska, M.~Henaff, M.~Mathieu, G.~B. Arous, and Y.~LeCun.
\newblock The loss surfaces of multilayer networks.
\newblock In {\em International Conference on Artificial Intelligence and
  Statistics}, 2015.

\bibitem{cohen2016group}
T.~Cohen and M.~Welling.
\newblock Group equivariant convolutional networks.
\newblock In {\em International Conference on Machine learning}, pages
  2990--2999, 2016.

\bibitem{cohen2019general}
T.~S. Cohen, M.~Geiger, and M.~Weiler.
\newblock A general theory of equivariant cnns on homogeneous spaces.
\newblock In {\em Advances in Neural Information Processing Systems}, pages
  9145--9156, 2019.

\bibitem{cuff2016differential}
P.~Cuff and L.~Yu.
\newblock Differential privacy as a mutual information constraint.
\newblock In {\em ACM SIGSAC Conference on Computer and Communications
  Security}, pages 43--54, 2016.

\bibitem{cullina2018pac}
D.~Cullina, A.~N. Bhagoji, and P.~Mittal.
\newblock {PAC}-learning in the presence of adversaries.
\newblock In {\em Advances in Neural Information Processing Systems}, pages
  230--241, 2018.

\bibitem{dai2018adversarial}
H.~Dai, H.~Li, T.~Tian, X.~Huang, L.~Wang, J.~Zhu, and L.~Song.
\newblock Adversarial attack on graph structured data.
\newblock {\em arXiv preprint arXiv:1806.02371}, 2018.

\bibitem{damianou2013deep}
A.~Damianou and N.~D. Lawrence.
\newblock Deep {Gaussian} processes.
\newblock In {\em Artificial intelligence and statistics}, pages 207--215,
  2013.

\bibitem{dechter1986learning}
R.~Dechter.
\newblock Learning while searching in constraint-satisfaction problems.
\newblock 1986.

\bibitem{deng2018deep}
L.~Deng and Y.~Liu.
\newblock {\em Deep learning in natural language processing}.
\newblock Springer, 2018.

\bibitem{deng2014deep}
L.~Deng and D.~Yu.
\newblock Deep learning: methods and applications.
\newblock {\em Foundations and Trends in Signal Processing}, 7(3--4):197--387,
  2014.

\bibitem{deng2020toward}
Z.~Deng, H.~He, and W.~J. Su.
\newblock Toward better generalization bounds with locally elastic stability.
\newblock {\em arXiv preprint arXiv:2010.13988}, 2020.

\bibitem{dieuleveut2016nonparametric}
A.~Dieuleveut and F.~Bach.
\newblock Nonparametric stochastic approximation with large step-sizes.
\newblock {\em The Annals of Statistics}, 44(4):1363--1399, 2016.

\bibitem{ding2014bayesian}
N.~Ding, Y.~Fang, R.~Babbush, C.~Chen, R.~D. Skeel, and H.~Neven.
\newblock {Bayesian} sampling using stochastic gradient thermostats.
\newblock In {\em Advances in Neural Information Processing Systems}, pages
  3203--3211, 2014.

\bibitem{dinh2017sharp}
L.~Dinh, R.~Pascanu, S.~Bengio, and Y.~Bengio.
\newblock Sharp minima can generalize for deep nets.
\newblock In {\em International Conference on Machine learning}, 2017.

\bibitem{pmlr-v80-draxler18a}
F.~Draxler, K.~Veschgini, M.~Salmhofer, and F.~Hamprecht.
\newblock Essentially no barriers in neural network energy landscape.
\newblock In {\em International Conference on Machine learning}, 2018.

\bibitem{du2019gradient}
S.~Du, J.~Lee, H.~Li, L.~Wang, and X.~Zhai.
\newblock Gradient descent finds global minima of deep neural networks.
\newblock In {\em International Conference on Machine learning}, pages
  1675--1685, 2019.

\bibitem{du2018many}
S.~S. Du, Y.~Wang, X.~Zhai, S.~Balakrishnan, R.~R. Salakhutdinov, and A.~Singh.
\newblock How many samples are needed to estimate a convolutional neural
  network?
\newblock In {\em Advances in Neural Information Processing Systems}, pages
  373--383, 2018.

\bibitem{du2018gradient1}
S.~S. Du, X.~Zhai, B.~Poczos, and A.~Singh.
\newblock Gradient descent provably optimizes over-parameterized neural
  networks.
\newblock In {\em International Conference on Learning Representations}, 2019.

\bibitem{duane1987hybrid}
S.~Duane, A.~D. Kennedy, B.~J. Pendleton, and D.~Roweth.
\newblock Hybrid {Monte Carlo}.
\newblock {\em Physics Letters B}, 195(2):216--222, 1987.

\bibitem{dudley1967sizes}
R.~M. Dudley.
\newblock The sizes of compact subsets of hilbert space and continuity of
  {Gaussian} processes.
\newblock {\em Journal of Functional Analysis}, 1(3):290--330, 1967.

\bibitem{duvenaud2014avoiding}
D.~Duvenaud, O.~Rippel, R.~Adams, and Z.~Ghahramani.
\newblock Avoiding pathologies in very deep networks.
\newblock In {\em Artificial Intelligence and Statistics}, pages 202--210,
  2014.

\bibitem{dwork2015preserving}
C.~Dwork, V.~Feldman, M.~Hardt, T.~Pitassi, O.~Reingold, and A.~L. Roth.
\newblock Preserving statistical validity in adaptive data analysis.
\newblock In {\em Annual ACM Symposium on Theory of Computing}, pages 117--126,
  2015.

\bibitem{dwork2012fairness}
C.~Dwork, M.~Hardt, T.~Pitassi, O.~Reingold, and R.~Zemel.
\newblock Fairness through awareness.
\newblock In {\em Innovations in Theoretical Computer Science Conference},
  pages 214--226, 2012.

\bibitem{dwork2013s}
C.~Dwork and D.~K. Mulligan.
\newblock It's not privacy, and it's not fair.
\newblock {\em Stanford Law Review Online}, 66:35, 2013.

\bibitem{dwork2014algorithmic}
C.~Dwork and A.~Roth.
\newblock The algorithmic foundations of differential privacy.
\newblock {\em Foundations and Trends{\textregistered} in Theoretical Computer
  Science}, 9(3--4):211--407, 2014.

\bibitem{dwork2016concentrated}
C.~Dwork and G.~N. Rothblum.
\newblock Concentrated differential privacy.
\newblock {\em arXiv preprint arXiv:1603.01887}, 2016.

\bibitem{dziugaite2020role}
G.~K. Dziugaite, K.~Hsu, W.~Gharbieh, and D.~M. Roy.
\newblock On the role of data in pac-bayes bounds.
\newblock {\em arXiv preprint arXiv:2006.10929}, 2020.

\bibitem{e2020towards}
W.~E, C.~Ma, S.~Wojtowytsch, and L.~Wu.
\newblock Towards a mathematical understanding of neural network-based machine
  learning: What we know and what we don't.
\newblock {\em arXiv preprint arXiv:2009.10713}, 2020.

\bibitem{faber2016machine}
F.~A. Faber, A.~Lindmaa, O.~A. Von~Lilienfeld, and R.~Armiento.
\newblock Machine learning energies of 2 million elpasolite (a b c 2 d 6)
  crystals.
\newblock {\em Physical Review Letters}, 117(13):135502, 2016.

\bibitem{faliagka2012application}
E.~Faliagka, K.~Ramantas, A.~Tsakalidis, and G.~Tzimas.
\newblock Application of machine learning algorithms to an online recruitment
  system.
\newblock In {\em International Conference on Internet and Web Applications and
  Services}, 2012.

\bibitem{feldman2015certifying}
M.~Feldman, S.~A. Friedler, J.~Moeller, C.~Scheidegger, and
  S.~Venkatasubramanian.
\newblock Certifying and removing disparate impact.
\newblock In {\em ACM SIGKDD International Conference on Knowledge Discovery
  and Data Mining}, pages 259--268, 2015.

\bibitem{garipov2018loss}
T.~Garipov, P.~Izmailov, D.~Podoprikhin, D.~P. Vetrov, and A.~G. Wilson.
\newblock Loss surfaces, mode connectivity, and fast ensembling of dnns.
\newblock In {\em Advances in Neural Information Processing Systems}, 2018.

\bibitem{gehring2017convolutional}
J.~Gehring, M.~Auli, D.~Grangier, D.~Yarats, and Y.~N. Dauphin.
\newblock Convolutional sequence to sequence learning.
\newblock In {\em International Conference on Machine learning}, pages
  1243--1252, 2017.

\bibitem{geman1984stochastic}
S.~Geman and D.~Geman.
\newblock Stochastic relaxation, {Gibbs} distributions, and the {Bayesian}
  restoration of images.
\newblock {\em IEEE Transactions on Pattern Analysis and Machine Intelligence},
  (6):721--741, 1984.

\bibitem{geumlek2017renyi}
J.~Geumlek, S.~Song, and K.~Chaudhuri.
\newblock Renyi differential privacy mechanisms for posterior sampling.
\newblock In {\em Advances in Neural Information Processing Systems}, pages
  5289--5298, 2017.

\bibitem{gilmer2018motivating}
J.~Gilmer, R.~P. Adams, I.~Goodfellow, D.~Andersen, and G.~E. Dahl.
\newblock Motivating the rules of the game for adversarial example research.
\newblock {\em arXiv preprint arXiv:1807.06732}, 2018.

\bibitem{goldberg1995bounding}
P.~W. Goldberg and M.~R. Jerrum.
\newblock Bounding the {Vapnik-Chervonenkis} dimension of concept classes
  parameterized by real numbers.
\newblock {\em Machine Learning}, 18(2-3):131--148, 1995.

\bibitem{Goldblum2020Truth}
M.~Goldblum, J.~Geiping, A.~Schwarzschild, M.~Moeller, and T.~Goldstein.
\newblock Truth or backpropaganda? {An} empirical investigation of deep
  learning theory.
\newblock In {\em International Conference on Learning Representations}, 2020.

\bibitem{golowich2017size}
N.~Golowich, A.~Rakhlin, and O.~Shamir.
\newblock Size-independent sample complexity of neural networks.
\newblock In {\em Annual Conference on Learning Theory}, pages 297--299, 2018.

\bibitem{goodfellow2014generative}
I.~Goodfellow, J.~Pouget-Abadie, M.~Mirza, B.~Xu, D.~Warde-Farley, S.~Ozair,
  A.~Courville, and Y.~Bengio.
\newblock Generative adversarial nets.
\newblock In {\em Advances in Neural Information Processing Systems}, pages
  2672--2680, 2014.

\bibitem{goodfellow2014explaining}
I.~J. Goodfellow, J.~Shlens, and C.~Szegedy.
\newblock Explaining and harnessing adversarial examples.
\newblock {\em arXiv preprint arXiv:1412.6572}, 2014.

\bibitem{goyal2017accurate}
P.~Goyal, P.~Doll{\'a}r, R.~Girshick, P.~Noordhuis, L.~Wesolowski, A.~Kyrola,
  A.~Tulloch, Y.~Jia, and K.~He.
\newblock Accurate, large minibatch sgd: training imagenet in 1 hour.
\newblock {\em arXiv preprint arXiv:1706.02677}, 2017.

\bibitem{graves2006connectionist}
A.~Graves, S.~Fern{\'a}ndez, F.~Gomez, and J.~Schmidhuber.
\newblock Connectionist temporal classification: labelling unsegmented sequence
  data with recurrent neural networks.
\newblock In {\em International Conference on Machine learning}, pages
  369--376, 2006.

\bibitem{graves2013speech}
A.~Graves, A.-r. Mohamed, and G.~Hinton.
\newblock Speech recognition with deep recurrent neural networks.
\newblock In {\em IEEE International Conference on Acoustics, Speech and Signal
  Processing}, pages 6645--6649, 2013.

\bibitem{guo2020deep}
Y.~Guo, H.~Wang, Q.~Hu, H.~Liu, L.~Liu, and M.~Bennamoun.
\newblock Deep learning for 3d point clouds: A survey.
\newblock {\em IEEE Transactions on Pattern Analysis and Machine Intelligence},
  2020.

\bibitem{haghifam2020sharpened}
M.~Haghifam, J.~Negrea, A.~Khisti, D.~M. Roy, and G.~K. Dziugaite.
\newblock Sharpened generalization bounds based on conditional mutual
  information and an application to noisy, iterative algorithms.
\newblock {\em arXiv preprint arXiv:2004.12983}, 2020.

\bibitem{hardt2016equality}
M.~Hardt, E.~Price, and N.~Srebro.
\newblock Equality of opportunity in supervised learning.
\newblock In {\em Advances in Neural Information Processing Systems}, pages
  3315--3323, 2016.

\bibitem{hardt2016train}
M.~Hardt, B.~Recht, and Y.~Singer.
\newblock Train faster, generalize better: Stability of stochastic gradient
  descent.
\newblock In {\em International Conference on Machine learning}, pages
  1225--1234, 2016.

\bibitem{harvey2017nearly}
N.~Harvey, C.~Liaw, and A.~Mehrabian.
\newblock Nearly-tight {VC}-dimension bounds for piecewise linear neural
  networks.
\newblock In {\em Annual Conference on Learning Theory}, pages 1064--1068,
  2017.

\bibitem{hastings1970monte}
W.~K. Hastings.
\newblock {Monte Carlo} sampling methods using {Markov} chains and their
  applications.
\newblock 1970.

\bibitem{haussler1995sphere}
D.~Haussler.
\newblock Sphere packing numbers for subsets of the boolean $n$-cube with
  bounded {Vapnik-Chervonenkis} dimension.
\newblock {\em Journal of Combinatorial Theory, Series A}, 69(2):217--232,
  1995.

\bibitem{hazan2015steps}
T.~Hazan and T.~Jaakkola.
\newblock Steps toward deep kernel methods from infinite neural networks.
\newblock {\em arXiv preprint arXiv:1508.05133}, 2015.

\bibitem{he2019control}
F.~He, T.~Liu, and D.~Tao.
\newblock Control batch size and learning rate to generalize well: Theoretical
  and empirical evidence.
\newblock In {\em Advances in Neural Information Processing Systems}, 2019.

\bibitem{he2020resnet}
F.~He, T.~Liu, and D.~Tao.
\newblock Why resnet works? residuals generalize.
\newblock {\em IEEE Transactions on Neural Networks and Learning Systems},
  2020.

\bibitem{he2020piecewise}
F.~He, B.~Wang, and D.~Tao.
\newblock Piecewise linear activations substantially shape the loss surfaces of
  neural networks.
\newblock In {\em International Conference on Learning Representations}, 2020.

\bibitem{he2020tighter}
F.~He, B.~Wang, and D.~Tao.
\newblock Tighter generalization bounds for iterative differentially private
  learning algorithms.
\newblock {\em arXiv preprint arXiv:2007.09371}, 2020.

\bibitem{he2020local}
H.~He and W.~Su.
\newblock The local elasticity of neural networks.
\newblock In {\em International Conference on Learning Representations}, 2020.

\bibitem{he2016deep}
K.~He, X.~Zhang, S.~Ren, and J.~Sun.
\newblock Deep residual learning for image recognition.
\newblock In {\em IEEE Conference on Computer Vision and Pattern Recognition},
  2016.

\bibitem{hebb1949organization}
D.~O. Hebb.
\newblock {\em The Organization of Behavior: A Neuropsychological Theory}.
\newblock J. Wiley; Chapman \& Hall, 1949.

\bibitem{hellstrom2020generalization}
F.~Hellstr{\"o}m and G.~Durisi.
\newblock Generalization bounds via information density and conditional
  information density.
\newblock {\em arXiv preprint arXiv:2005.08044}, 2020.

\bibitem{hensman2014nested}
J.~Hensman and N.~D. Lawrence.
\newblock Nested variational compression in deep {Gaussian} processes.
\newblock {\em arXiv preprint arXiv:1412.1370}, 2014.

\bibitem{hinton2006fast}
G.~E. Hinton, S.~Osindero, and Y.-W. Teh.
\newblock A fast learning algorithm for deep belief nets.
\newblock {\em Neural Computation}, 18(7):1527--1554, 2006.

\bibitem{hoffman2013stochastic}
M.~D. Hoffman, D.~M. Blei, C.~Wang, and J.~Paisley.
\newblock Stochastic variational inference.
\newblock {\em Journal of Machine Learning Research}, 14(1):1303--1347, 2013.

\bibitem{jacot2018neural}
A.~Jacot, F.~Gabriel, and C.~Hongler.
\newblock Neural tangent kernel: Convergence and generalization in neural
  networks.
\newblock In {\em Advances in Neural Information Processing Systems}, pages
  8571--8580, 2018.

\bibitem{jastrzkebski2017three}
S.~Jastrz{\k{e}}bski, Z.~Kenton, D.~Arpit, N.~Ballas, A.~Fischer, Y.~Bengio,
  and A.~Storkey.
\newblock Three factors influencing minima in sgd.
\newblock {\em arXiv preprint arXiv:1711.04623}, 2017.

\bibitem{jia2017improving}
K.~Jia, D.~Tao, S.~Gao, and X.~Xu.
\newblock Improving training of deep neural networks via singular value
  bounding.
\newblock In {\em IEEE Conference on Computer Vision and Pattern Recognition},
  pages 4344--4352, 2017.

\bibitem{kamiran2012data}
F.~Kamiran and T.~Calders.
\newblock Data preprocessing techniques for classification without
  discrimination.
\newblock {\em Knowledge and Information Systems}, 33(1):1--33, 2012.

\bibitem{kamishima2012fairness}
T.~Kamishima, S.~Akaho, H.~Asoh, and J.~Sakuma.
\newblock Fairness-aware classifier with prejudice remover regularizer.
\newblock In {\em Joint European Conference on Machine Learning and Knowledge
  Discovery in Databases}, pages 35--50, 2012.

\bibitem{kawaguchi2016deep}
K.~Kawaguchi.
\newblock Deep learning without poor local minima.
\newblock In {\em Advances in Neural Information Processing Systems}, 2016.

\bibitem{kawaguchi2017generalization}
K.~Kawaguchi, L.~P. Kaelbling, and Y.~Bengio.
\newblock Generalization in deep learning.
\newblock {\em arXiv:1710.05468}, 2017.

\bibitem{kearns2018preventing}
M.~Kearns, S.~Neel, A.~Roth, and Z.~S. Wu.
\newblock Preventing fairness gerrymandering: Auditing and learning for
  subgroup fairness.
\newblock In {\em International Conference on Machine learning}, pages
  2564--2572, 2018.

\bibitem{keskar2016large}
N.~S. Keskar, D.~Mudigere, J.~Nocedal, M.~Smelyanskiy, and P.~T.~P. Tang.
\newblock On large-batch training for deep learning: Generalization gap and
  sharp minima.
\newblock In {\em International Conference on Leanring Representations}, 2017.

\bibitem{khandani2010consumer}
A.~E. Khandani, A.~J. Kim, and A.~W. Lo.
\newblock Consumer credit-risk models via machine-learning algorithms.
\newblock {\em Journal of Banking \& Finance}, 34(11):2767--2787, 2010.

\bibitem{khim2018adversarial}
J.~Khim and P.-L. Loh.
\newblock Adversarial risk bounds via function transformation.
\newblock {\em arXiv preprint arXiv:1810.09519}, 2018.

\bibitem{kingma2014auto}
D.~P. Kingma and M.~Welling.
\newblock Auto-encoding variational {Bayes}.
\newblock In {\em International Conference on Learning Representations}, 2014.

\bibitem{kobyzev2020normalizing}
I.~Kobyzev, S.~Prince, and M.~Brubaker.
\newblock Normalizing flows: An introduction and review of current methods.
\newblock {\em IEEE Transactions on Pattern Analysis and Machine Intelligence},
  2020.

\bibitem{koltchinskii2001rademacher}
V.~Koltchinskii.
\newblock Rademacher penalties and structural risk minimization.
\newblock {\em IEEE Transactions on Information Theory}, 47(5):1902--1914,
  2001.

\bibitem{koltchinskii2000rademacher}
V.~Koltchinskii and D.~Panchenko.
\newblock Rademacher processes and bounding the risk of function learning.
\newblock In {\em High Dimensional Probability II}, pages 443--457. Springer,
  2000.

\bibitem{koltchinskii2002empirical}
V.~Koltchinskii and D.~Panchenko.
\newblock Empirical margin distributions and bounding the generalization error
  of combined classifiers.
\newblock {\em The Annals of Statistics}, 30(1):1--50, 2002.

\bibitem{krizhevsky2012imagenet}
A.~Krizhevsky, I.~Sutskever, and G.~E. Hinton.
\newblock Imagenet classification with deep convolutional neural networks.
\newblock In {\em Advances in Neural Information Processing Systems}, pages
  1097--1105, 2012.

\bibitem{krueger2017deep}
D.~Krueger, N.~Ballas, S.~Jastrzebski, D.~Arpit, M.~S. Kanwal, T.~Maharaj,
  E.~Bengio, A.~Fischer, and A.~Courville.
\newblock Deep nets don't learn via memorization.
\newblock 2017.

\bibitem{kuditipudi2019explaining}
R.~Kuditipudi, X.~Wang, H.~Lee, Y.~Zhang, Z.~Li, W.~Hu, S.~Arora, and R.~Ge.
\newblock Explaining landscape connectivity of low-cost solutions for
  multilayer nets.
\newblock In {\em Advances in Neural Information Processing Systems}, 2019.

\bibitem{kulikowski1980artificial}
C.~A. Kulikowski.
\newblock Artificial intelligence methods and systems for medical consultation.
\newblock {\em IEEE Transactions on Pattern Analysis and Machine Intelligence},
  (5):464--476, 1980.

\bibitem{larochelle2012neural}
H.~Larochelle and S.~Lauly.
\newblock A neural autoregressive topic model.
\newblock In {\em Advances in Neural Information Processing Systems}, pages
  2708--2716, 2012.

\bibitem{lawrence2007hierarchical}
N.~D. Lawrence and A.~J. Moore.
\newblock Hierarchical {Gaussian} process latent variable models.
\newblock In {\em International Conference on Machine learning}, pages
  481--488, 2007.

\bibitem{lecun2015deep}
Y.~LeCun, Y.~Bengio, and G.~Hinton.
\newblock Deep learning.
\newblock {\em Nature}, 521(7553):436, 2015.

\bibitem{lecun1998gradient}
Y.~LeCun, L.~Bottou, Y.~Bengio, and P.~Haffner.
\newblock Gradient-based learning applied to document recognition.
\newblock {\em Proceedings of the IEEE}, 86(11):2278--2324, 1998.

\bibitem{lee2018deep}
J.~Lee, Y.~Bahri, R.~Novak, S.~S. Schoenholz, J.~Pennington, and
  J.~Sohl-Dickstein.
\newblock Deep neural networks as {Gaussian} processes.
\newblock In {\em International Conference on Learning Representations}, 2018.

\bibitem{lee2019wide}
J.~Lee, L.~Xiao, S.~Schoenholz, Y.~Bahri, R.~Novak, J.~Sohl-Dickstein, and
  J.~Pennington.
\newblock Wide neural networks of any depth evolve as linear models under
  gradient descent.
\newblock In {\em Advances in Neural Information Processing Systems}, pages
  8572--8583, 2019.

\bibitem{lever2013tighter}
G.~Lever, F.~Laviolette, and J.~Shawe-Taylor.
\newblock Tighter pac-bayes bounds through distribution-dependent priors.
\newblock {\em Theoretical Computer Science}, 473:4--28, 2013.

\bibitem{li2018second}
B.~Li, C.~Chen, W.~Wang, and L.~Carin.
\newblock Second-order adversarial attack and certifiable robustness.
\newblock 2018.

\bibitem{li2018over}
D.~Li, T.~Ding, and R.~Sun.
\newblock Over-parameterized deep neural networks have no strict local minima
  for any continuous activations.
\newblock {\em arXiv preprint arXiv:1812.11039}, 2018.

\bibitem{li2019generalization}
J.~Li, X.~Luo, and M.~Qiao.
\newblock On generalization error bounds of noisy gradient methods for
  non-convex learning.
\newblock {\em arXiv preprint arXiv:1902.00621}, 2019.

\bibitem{li2018pointcnn}
Y.~Li, R.~Bu, M.~Sun, W.~Wu, X.~Di, and B.~Chen.
\newblock Pointcnn: Convolution on x-transformed points.
\newblock In {\em Advances in Neural Information Processing Systems}, pages
  820--830, 2018.

\bibitem{li2017convergence}
Y.~Li and Y.~Yuan.
\newblock Convergence analysis of two-layer neural networks with {ReLU}
  activation.
\newblock In {\em Advances in Neural Information Processing Systems}, pages
  597--607, 2017.

\bibitem{liao2017hypothesis}
J.~Liao, L.~Sankar, V.~Y. Tan, and F.~du~Pin~Calmon.
\newblock Hypothesis testing under mutual information privacy constraints in
  the high privacy regime.
\newblock {\em IEEE Transactions on Information Forensics and Security},
  13(4):1058--1071, 2017.

\bibitem{lin2016generalization}
J.~Lin, R.~Camoriano, and L.~Rosasco.
\newblock Generalization properties and implicit regularization for multiple
  passes {SGM}.
\newblock In {\em International Conference on Machine learning}, pages
  2340--2348, 2016.

\bibitem{lin2016optimal}
J.~Lin and L.~Rosasco.
\newblock Optimal learning for multi-pass stochastic gradient methods.
\newblock In {\em Advances in Neural Information Processing Systems}, pages
  4556--4564, 2016.

\bibitem{lin2019generalization}
S.~Lin and J.~Zhang.
\newblock Generalization bounds for convolutional neural networks.
\newblock {\em arXiv preprint arXiv:1910.01487}, 2019.

\bibitem{litjens2017survey}
G.~Litjens, T.~Kooi, B.~E. Bejnordi, A.~A.~A. Setio, F.~Ciompi, M.~Ghafoorian,
  J.~A. Van Der~Laak, B.~Van~Ginneken, and C.~I. S{\'a}nchez.
\newblock A survey on deep learning in medical image analysis.
\newblock {\em Medical Image Analysis}, 42:60--88, 2017.

\bibitem{lohr2018facial}
S.~Lohr.
\newblock Facial recognition is accurate, if you’re a white guy.
\newblock {\em New York Times}, 9, 2018.

\bibitem{louizos2015variational}
C.~Louizos, K.~Swersky, Y.~Li, M.~Welling, and R.~Zemel.
\newblock The variational fair autoencoder.
\newblock {\em arXiv preprint arXiv:1511.00830}, 2015.

\bibitem{louizos2017multiplicative}
C.~Louizos and M.~Welling.
\newblock Multiplicative normalizing flows for variational {Bayesian} neural
  networks.
\newblock In {\em International Conference on Machine learning}, pages
  2218--2227, 2017.

\bibitem{lu2017depth}
H.~Lu and K.~Kawaguchi.
\newblock Depth creates no bad local minima.
\newblock {\em arXiv preprint arXiv:1702.08580}, 2017.

\bibitem{ma2015complete}
Y.-A. Ma, T.~Chen, and E.~Fox.
\newblock A complete recipe for stochastic gradient mcmc.
\newblock In {\em Advances in Neural Information Processing Systems}, pages
  2917--2925, 2015.

\bibitem{maass1997bounds}
W.~Maass.
\newblock Bounds for the computational power and learning complexity of analog
  neural nets.
\newblock {\em SIAM Journal on Computing}, 26(3):708--732, 1997.

\bibitem{mandt2017stochastic}
S.~Mandt, M.~D. Hoffman, and D.~M. Blei.
\newblock Stochastic gradient descent as approximate {Bayesian} inference.
\newblock {\em Journal of Machine Learning Research}, 18(1):4873--4907, 2017.

\bibitem{mcallester1999pac}
D.~A. McAllester.
\newblock {PAC-Bayesian} model averaging.
\newblock In {\em Annual Conference of Learning Theory}, volume~99, pages
  164--170, 1999.

\bibitem{mcallester1999some}
D.~A. McAllester.
\newblock Some {PAC-Bayesian} theorems.
\newblock {\em Machine Learning}, 37(3):355--363, 1999.

\bibitem{mcculloch1943logical}
W.~S. McCulloch and W.~Pitts.
\newblock A logical calculus of the ideas immanent in nervous activity.
\newblock {\em The Bulletin of Mathematical Biophysics}, 5(4):115--133, 1943.

\bibitem{mei2018landscape}
S.~Mei, Y.~Bai, and A.~Montanari.
\newblock The landscape of empirical risk for nonconvex losses.
\newblock {\em The Annals of Statistics}, 46(6A):2747--2774, 2018.

\bibitem{menon2018cost}
A.~K. Menon and R.~C. Williamson.
\newblock The cost of fairness in binary classification.
\newblock In {\em Conference on Fairness, Accountability and Transparency},
  pages 107--118, 2018.

\bibitem{merity2018regularizing}
S.~Merity, N.~S. Keskar, and R.~Socher.
\newblock Regularizing and optimizing lstm language models.
\newblock In {\em International Conference on Learning Representations}, 2018.

\bibitem{min2020curious}
Y.~Min, L.~Chen, and A.~Karbasi.
\newblock The curious case of adversarially robust models: More data can help,
  double descend, or hurt generalization.
\newblock {\em arXiv preprint arXiv:2002.11080}, 2020.

\bibitem{mironov2017renyi}
I.~Mironov.
\newblock R{\'e}nyi differential privacy.
\newblock In {\em IEEE Computer Security Foundations Symposium}, pages
  263--275, 2017.

\bibitem{miyato2018spectral}
T.~Miyato, T.~Kataoka, M.~Koyama, and Y.~Yoshida.
\newblock Spectral normalization for generative adversarial networks.
\newblock In {\em International Conference on Learning Representations}, 2018.

\bibitem{mohri2018foundations}
M.~Mohri, A.~Rostamizadeh, and A.~Talwalkar.
\newblock {\em Foundations of machine learning}.
\newblock MIT press, 2018.

\bibitem{montasser2019vc}
O.~Montasser, S.~Hanneke, and N.~Srebro.
\newblock {VC} classes are adversarially robustly learnable, but only
  improperly.
\newblock In {\em Conference on Learning Theory}, pages 2512--2530, 2019.

\bibitem{mou2017generalization}
W.~Mou, L.~Wang, X.~Zhai, and K.~Zheng.
\newblock Generalization bounds of sgld for non-convex learning: Two
  theoretical viewpoints.
\newblock In {\em Annual Conference On Learning Theory}, 2018.

\bibitem{musavi1994generalization}
M.~T. Musavi, K.~H. Chan, D.~M. Hummels, and K.~Kalantri.
\newblock On the generalization ability of neural network classifiers.
\newblock {\em IEEE Transactions on Pattern Analysis and Machine Intelligence},
  16(6):659--663, 1994.

\bibitem{neal1995bayesian}
R.~M. Neal.
\newblock {\em {Bayesian} LEARNING FOR NEURAL NETWORKS}.
\newblock PhD thesis, University of Toronto, 1995.

\bibitem{neal1996priors}
R.~M. Neal.
\newblock Priors for infinite networks.
\newblock In {\em Bayesian Learning for Neural Networks}, pages 29--53.
  Springer, 1996.

\bibitem{negrea2019information}
J.~Negrea, M.~Haghifam, G.~K. Dziugaite, A.~Khisti, and D.~M. Roy.
\newblock Information-theoretic generalization bounds for sgld via
  data-dependent estimates.
\newblock In {\em Advances in Neural Information Processing Systems}, pages
  11015--11025, 2019.

\bibitem{neyshabur2017pac}
B.~Neyshabur, S.~Bhojanapalli, and N.~Srebro.
\newblock A {PAC-Bayesian} approach to spectrally-normalized margin bounds for
  neural networks.
\newblock {\em arXiv preprint arXiv:1707.09564}, 2017.

\bibitem{neyshabur2015norm}
B.~Neyshabur, R.~Tomioka, and N.~Srebro.
\newblock Norm-based capacity control in neural networks.
\newblock In {\em Conference on Learning Theory}, pages 1376--1401, 2015.

\bibitem{nguyen2015deep}
A.~Nguyen, J.~Yosinski, and J.~Clune.
\newblock Deep neural networks are easily fooled: High confidence predictions
  for unrecognizable images.
\newblock In {\em IEEE Conference on Computer Vision and Pattern Recognition},
  pages 427--436, 2015.

\bibitem{nguyen2019on}
Q.~Nguyen.
\newblock On connected sublevel sets in deep learning.
\newblock In {\em International Conference on Machine learning}, 2019.

\bibitem{nguyen2018on}
Q.~Nguyen, M.~C. Mukkamala, and M.~Hein.
\newblock On the loss landscape of a class of deep neural networks with no bad
  local valleys.
\newblock In {\em International Conference on Learning Representations}, 2019.

\bibitem{nissim2015generalization}
K.~Nissim and U.~Stemmer.
\newblock On the generalization properties of differential privacy.
\newblock {\em CoRR, abs/1504.05800}, 2015.

\bibitem{ntampaka2016dynamical}
M.~Ntampaka, H.~Trac, D.~J. Sutherland, S.~Fromenteau, B.~P{\'o}czos, and
  J.~Schneider.
\newblock Dynamical mass measurements of contaminated galaxy clusters using
  machine learning.
\newblock {\em The Astrophysical Journal}, 831(2):135, 2016.

\bibitem{oneto2017differential}
L.~Oneto, S.~Ridella, and D.~Anguita.
\newblock Differential privacy and generalization: Sharper bounds with
  applications.
\newblock {\em Pattern Recognition Letters}, 89:31--38, 2017.

\bibitem{otter2020survey}
D.~W. Otter, J.~R. Medina, and J.~K. Kalita.
\newblock A survey of the usages of deep learning for natural language
  processing.
\newblock {\em IEEE Transactions on Neural Networks and Learning Systems},
  2020.

\bibitem{papernot2017practical}
N.~Papernot, P.~McDaniel, I.~Goodfellow, S.~Jha, Z.~B. Celik, and A.~Swami.
\newblock Practical black-box attacks against machine learning.
\newblock In {\em ACM on Asia Conference on Computer and Communications
  Security}, pages 506--519, 2017.

\bibitem{papernot2016limitations}
N.~Papernot, P.~McDaniel, S.~Jha, M.~Fredrikson, Z.~B. Celik, and A.~Swami.
\newblock The limitations of deep learning in adversarial settings.
\newblock In {\em IEEE European Symposium on Security and Privacy}, pages
  372--387, 2016.

\bibitem{papernot2016towards}
N.~Papernot, P.~McDaniel, A.~Sinha, and M.~Wellman.
\newblock Towards the science of security and privacy in machine learning.
\newblock {\em arXiv preprint arXiv:1611.03814}, 2016.

\bibitem{papyan2018full}
V.~Papyan.
\newblock The full spectrum of deep net hessians at scale: Dynamics with sample
  size.
\newblock {\em arXiv preprint arXiv:1811.07062}, 2018.

\bibitem{patterson2013stochastic}
S.~Patterson and Y.~W. Teh.
\newblock Stochastic gradient {Riemannian} {Langevin} dynamics on the
  probability simplex.
\newblock In {\em Advances in Neural Information Processing Systems}, pages
  3102--3110, 2013.

\bibitem{pensia2018generalization}
A.~Pensia, V.~Jog, and P.-L. Loh.
\newblock Generalization error bounds for noisy, iterative algorithms.
\newblock In {\em {IEEE} International Symposium on Information Theory}, 2018.

\bibitem{peters2018deep}
M.~E. Peters, M.~Neumann, M.~Iyyer, M.~Gardner, C.~Clark, K.~Lee, and
  L.~Zettlemoyer.
\newblock Deep contextualized word representations.
\newblock In {\em Annual Conference of the North American Chapter of the
  Association for Computational Linguistics}, pages 2227--2237, 2018.

\bibitem{pydi2020adversarial}
M.~S. Pydi and V.~Jog.
\newblock Adversarial risk via optimal transport and optimal couplings.
\newblock In {\em International Conference on Machine learning}, pages
  7814--7823, 2020.

\bibitem{raghunathan2018certified}
A.~Raghunathan, J.~Steinhardt, and P.~Liang.
\newblock Certified defenses against adversarial examples.
\newblock In {\em International Conference on Learning Representations}, 2018.

\bibitem{raginsky2017non}
M.~Raginsky, A.~Rakhlin, and M.~Telgarsky.
\newblock Non-convex learning via stochastic gradient {Langevin} dynamics: A
  nonasymptotic analysis.
\newblock In {\em Conference on Learning Theory}, pages 1674--1703, 2017.

\bibitem{ramadorai2017predictably}
T.~Ramadorai, A.~Fuster, P.~Goldsmith-Pinkham, and A.~Walther.
\newblock Predictably unequal? the effects of machine learning on credit
  markets.
\newblock 2017.

\bibitem{ranzato2006efficient}
M.~Ranzato, C.~Poultney, S.~Chopra, and Y.~Cun.
\newblock Efficient learning of sparse representations with an energy-based
  model.
\newblock {\em Advances in Neural Information Processing Systems},
  19:1137--1144, 2006.

\bibitem{ravanbakhsh2016estimating}
S.~Ravanbakhsh, J.~Oliva, S.~Fromenteau, L.~Price, S.~Ho, J.~Schneider, and
  B.~P{\'o}czos.
\newblock Estimating cosmological parameters from the dark matter distribution.
\newblock In {\em International Conference on Machine learning}, pages
  2407--2416, 2016.

\bibitem{robbins1951stochastic}
H.~Robbins and S.~Monro.
\newblock A stochastic approximation method.
\newblock {\em The Annals of Mathematical Statistics}, pages 400--407, 1951.

\bibitem{rosenblatt1958perceptron}
F.~Rosenblatt.
\newblock The perceptron: a probabilistic model for information storage and
  organization in the brain.
\newblock {\em Psychological Review}, 65(6):386, 1958.

\bibitem{roth2018bayesian}
W.~Roth and F.~Pernkopf.
\newblock {Bayesian} neural networks with weight sharing using {Dirichlet}
  processes.
\newblock {\em IEEE Transactions on Pattern Analysis and Machine Intelligence},
  42(1):246--252, 2018.

\bibitem{rumelhart1986learning}
D.~E. Rumelhart, G.~E. Hinton, and R.~J. Williams.
\newblock Learning representations by back-propagating errors.
\newblock {\em Nature}, 323(6088):533--536, 1986.

\bibitem{safran2017spurious}
I.~Safran and O.~Shamir.
\newblock Spurious local minima are common in two-layer {ReLU} neural networks.
\newblock In {\em International Conference on Machine learning}, 2018.

\bibitem{sagun2016singularity}
L.~Sagun, L.~Bottou, and Y.~LeCun.
\newblock Singularity of the hessian in deep learning.
\newblock {\em arXiv preprint arXiv:1611.07476}, 2016.

\bibitem{sagun2018empirical}
L.~Sagun, U.~Evci, V.~U. Guney, Y.~Dauphin, and L.~Bottou.
\newblock Empirical analysis of the hessian of over-parametrized neural
  networks.
\newblock In {\em International Conference on Learning Representations
  Workshop}, 2018.

\bibitem{sannai2019improved}
A.~Sannai and M.~Imaizumi.
\newblock Improved generalization bound of permutation invariant deep neural
  networks.
\newblock {\em arXiv preprint arXiv:1910.06552}, 2019.

\bibitem{sannai2019universal}
A.~Sannai, Y.~Takai, and M.~Cordonnier.
\newblock Universal approximations of permutation invariant/equivariant
  functions by deep neural networks.
\newblock {\em arXiv preprint arXiv:1903.01939}, 2019.

\bibitem{schapire1998boosting}
R.~E. Schapire, Y.~Freund, P.~Bartlett, and W.~S. Lee.
\newblock Boosting the margin: A new explanation for the effectiveness of
  voting methods.
\newblock {\em The Annals of Statistics}, 26(5):1651--1686, 1998.

\bibitem{schmidt2018adversarially}
L.~Schmidt, S.~Santurkar, D.~Tsipras, K.~Talwar, and A.~Madry.
\newblock Adversarially robust generalization requires more data.
\newblock {\em Advances in Neural Information Processing Systems},
  31:5014--5026, 2018.

\bibitem{silver2016mastering}
D.~Silver, A.~Huang, C.~J. Maddison, A.~Guez, L.~Sifre, G.~Van Den~Driessche,
  J.~Schrittwieser, I.~Antonoglou, V.~Panneershelvam, and M.~Lanctot.
\newblock Mastering the game of go with deep neural networks and tree search.
\newblock {\em Nature}, 529(7587):484, 2016.

\bibitem{smith2017don}
S.~L. Smith, P.-J. Kindermans, C.~Ying, and Q.~V. Le.
\newblock Don't decay the learning rate, increase the batch size.
\newblock In {\em International Conference on Learning Representations}, 2018.

\bibitem{smith2018bayesian}
S.~L. Smith and Q.~V. Le.
\newblock A {Bayesian} perspective on generalization and stochastic gradient
  descent.
\newblock In {\em International Conference on Learning Representations}, 2018.

\bibitem{soudry2018exponentially}
D.~Soudry and E.~Hoffer.
\newblock Exponentially vanishing sub-optimal local minima in multilayer neural
  networks.
\newblock In {\em International Conference on Learning Representations
  Workshop}, 2018.

\bibitem{steinke2020reasoning}
T.~Steinke and L.~Zakynthinou.
\newblock Reasoning about generalization via conditional mutual information.
\newblock {\em arXiv preprint arXiv:2001.09122}, 2020.

\bibitem{sun2019optimization}
R.~Sun.
\newblock Optimization for deep learning: Theory and algorithms.
\newblock {\em arXiv preprint arXiv:1912.08957}, 2019.

\bibitem{sun2006road}
Z.~Sun, G.~Bebis, and R.~Miller.
\newblock On-road vehicle detection: A review.
\newblock {\em IEEE Transactions on Pattern Analysis and Machine Intelligence},
  28(5):694--711, 2006.

\bibitem{sutskever2014sequence}
I.~Sutskever, O.~Vinyals, and Q.~V. Le.
\newblock Sequence to sequence learning with neural networks.
\newblock In {\em Advances in Neural Information Processing Systems}, pages
  3104--3112, 2014.

\bibitem{swirszcz2016local}
G.~Swirszcz, W.~M. Czarnecki, and R.~Pascanu.
\newblock Local minima in training of deep networks.
\newblock {\em arXiv preprint arXiv:1611:06310}, 2016.

\bibitem{szegedy2014intriguing}
C.~Szegedy, W.~Zaremba, I.~Sutskever, J.~Bruna, D.~Erhan, I.~Goodfellow, and
  R.~Fergus.
\newblock Intriguing properties of neural networks.
\newblock In {\em International Conference on Learning Representations}, 2014.

\bibitem{taskar2004max}
B.~Taskar, C.~Guestrin, and D.~Koller.
\newblock Max-margin {Markov} networks.
\newblock In {\em Advances in Neural Information Processing Systems}, pages
  25--32, 2004.

\bibitem{tu2020understanding}
Z.~Tu, F.~He, and D.~Tao.
\newblock Understanding generalization in recurrent neural networks.
\newblock In {\em International Conference on Learning Representations}, 2020.

\bibitem{tu2019theoretical}
Z.~Tu, J.~Zhang, and D.~Tao.
\newblock Theoretical analysis of adversarial learning: A minimax approach.
\newblock In {\em Advances in Neural Information Processing Systems}, pages
  12280--12290, 2019.

\bibitem{tzen2018local}
B.~Tzen, T.~Liang, and M.~Raginsky.
\newblock Local optimality and generalization guarantees for the {Langevin}
  algorithm via empirical metastability.
\newblock In {\em Conference On Learning Theory}, pages 857--875, 2018.

\bibitem{vapnik2006estimation}
V.~Vapnik.
\newblock {\em Estimation of Dependences based on Empirical Data}.
\newblock Springer Science \& Business Media, 2006.

\bibitem{vapnik2013nature}
V.~Vapnik.
\newblock {\em The Nature of Statistical Learning Theory}.
\newblock Springer Science \& Business Media, 2013.

\bibitem{wang2016relation}
W.~Wang, L.~Ying, and J.~Zhang.
\newblock On the relation between identifiability, differential privacy, and
  mutual-information privacy.
\newblock {\em IEEE Transactions on Information Theory}, 62(9):5018--5029,
  2016.

\bibitem{wei2019regularization}
C.~Wei, J.~D. Lee, Q.~Liu, and T.~Ma.
\newblock Regularization matters: Generalization and optimization of neural
  nets vs their induced kernel.
\newblock In {\em Advances in Neural Information Processing Systems}, pages
  9712--9724, 2019.

\bibitem{wei2017early}
Y.~Wei, F.~Yang, and M.~J. Wainwright.
\newblock Early stopping for kernel boosting algorithms: A general analysis
  with localized complexities.
\newblock In {\em Advances in Neural Information Processing Systems}, pages
  6065--6075, 2017.

\bibitem{weinan2017proposal}
E.~Weinan.
\newblock A proposal on machine learning via dynamical systems.
\newblock {\em Communications in Mathematics and Statistics}, 5(1):1--11, 2017.

\bibitem{welling2011bayesian}
M.~Welling and Y.~W. Teh.
\newblock {Bayesian} learning via stochastic gradient {Langevin} dynamics.
\newblock In {\em International Conference on Machine learning}, pages
  681--688, 2011.

\bibitem{wen2019interplay}
Y.~Wen, K.~Luk, M.~Gazeau, G.~Zhang, H.~Chan, and J.~Ba.
\newblock Interplay between optimization and generalization of stochastic
  gradient descent with covariance noise.
\newblock {\em arXiv preprint arXiv:1902.08234}, 2019.

\bibitem{williams1996computing}
C.~Williams.
\newblock Computing with infinite networks.
\newblock {\em Advances in Neural Information Processing Systems}, 9:295--301,
  1996.

\bibitem{witten2016data}
I.~H. Witten, E.~Frank, M.~A. Hall, and C.~J. Pal.
\newblock {\em Data Mining: Practical machine learning tools and techniques}.
\newblock Morgan Kaufmann, 2016.

\bibitem{woodworth2017learning}
B.~Woodworth, S.~Gunasekar, M.~I. Ohannessian, and N.~Srebro.
\newblock Learning non-discriminatory predictors.
\newblock {\em arXiv preprint arXiv:1702.06081}, 2017.

\bibitem{Xiang:EECS-2020-145}
C.~Xiang.
\newblock {\em The Interplay between Sampling and Optimization}.
\newblock PhD thesis, EECS Department, University of California, Berkeley,
  2020.

\bibitem{xu2017information}
A.~Xu and M.~Raginsky.
\newblock Information-theoretic analysis of generalization capability of
  learning algorithms.
\newblock In {\em Advances in Neural Information Processing Systems}, pages
  2524--2533, 2017.

\bibitem{xu2011sparse}
H.~Xu, C.~Caramanis, and S.~Mannor.
\newblock Sparse algorithms are not stable: A no-free-lunch theorem.
\newblock {\em IEEE Transactions on Pattern Analysis and Machine Intelligence},
  34(1):187--193, 2011.

\bibitem{yin2019rademacher}
D.~Yin, R.~Kannan, and P.~Bartlett.
\newblock Rademacher complexity for adversarially robust generalization.
\newblock In {\em International Conference on Machine learning}, pages
  7085--7094, 2019.

\bibitem{ying2008online}
Y.~Ying and M.~Pontil.
\newblock Online gradient descent learning algorithms.
\newblock {\em Foundations of Computational Mathematics}, 8(5):561--596, 2008.

\bibitem{yu2018qanet}
A.~W. Yu, D.~Dohan, M.-T. Luong, R.~Zhao, K.~Chen, M.~Norouzi, and Q.~V. Le.
\newblock Qanet: Combining local convolution with global self-attention for
  reading comprehension.
\newblock In {\em International Conference on Learning Representations}, 2018.

\bibitem{yun2018small}
C.~Yun, S.~Sra, and A.~Jadbabaie.
\newblock Small nonlinearities in activation functions create bad local minima
  in neural networks.
\newblock In {\em International Conference on Learning Representations}, 2019.

\bibitem{zafar2017fairness}
M.~B. Zafar, I.~Valera, M.~G. Rogriguez, and K.~P. Gummadi.
\newblock Fairness constraints: Mechanisms for fair classification.
\newblock In {\em Artificial Intelligence and Statistics}, pages 962--970,
  2017.

\bibitem{zaheer2017deep}
M.~Zaheer, S.~Kottur, S.~Ravanbhakhsh, B.~Póczos, R.~Salakhutdinov, and A.~J.
  Smola.
\newblock Deep sets.
\newblock In {\em Advances in Neural Information Processing Systems}, 2017.

\bibitem{zemel2013learning}
R.~Zemel, Y.~Wu, K.~Swersky, T.~Pitassi, and C.~Dwork.
\newblock Learning fair representations.
\newblock In {\em International Conference on Machine Learning}, pages
  325--333, 2013.

\bibitem{zhang2017understanding}
C.~Zhang, S.~Bengio, M.~Hardt, B.~Recht, and O.~Vinyals.
\newblock Understanding deep learning requires rethinking generalization.
\newblock In {\em International Conference on Learning Representations}, 2017.

\bibitem{zhang2018advances}
C.~Zhang, J.~B{\"u}tepage, H.~Kjellstr{\"o}m, and S.~Mandt.
\newblock Advances in variational inference.
\newblock {\em IEEE Transactions on Pattern Analysis and Machine Intelligence},
  41(8):2008--2026, 2018.

\bibitem{zhang2020deep}
H.~Zhang, B.~Chen, Y.~Cong, D.~Guo, H.~Liu, and M.~Zhou.
\newblock Deep autoencoding topic model with scalable hybrid {Bayesian}
  inference.
\newblock {\em IEEE Transactions on Pattern Analysis and Machine Intelligence},
  2020.

\bibitem{zhang2019deep}
H.~Zhang, J.~Shao, and R.~Salakhutdinov.
\newblock Deep neural networks with multi-branch architectures are
  intrinsically less non-convex.
\newblock In {\em International Conference on Artificial Intelligence and
  Statistics}, 2019.

\bibitem{zhang2019theoretically}
H.~Zhang, Y.~Yu, J.~Jiao, E.~P. Xing, L.~E. Ghaoui, and M.~I. Jordan.
\newblock Theoretically principled trade-off between robustness and accuracy.
\newblock {\em arXiv preprint arXiv:1901.08573}, 2019.

\bibitem{zhang2018information}
J.~Zhang, T.~Liu, and D.~Tao.
\newblock An information-theoretic view for deep learning.
\newblock {\em arXiv preprint arXiv:1804.09060}, 2018.

\bibitem{zheng2019distributionally}
T.~Zheng, C.~Chen, and K.~Ren.
\newblock Distributionally adversarial attack.
\newblock In {\em AAAI Conference on Artificial Intelligence}, volume~33, pages
  2253--2260, 2019.

\bibitem{zhou2017landscape}
P.~Zhou and J.~Feng.
\newblock Empirical risk landscape analysis for understanding deep neural
  networks.
\newblock In {\em International Conference on Learning Representations}, 2018.

\bibitem{zhou2018understanding}
P.~Zhou and J.~Feng.
\newblock Understanding generalization and optimization performance of deep
  cnns.
\newblock In {\em International Conference on Machine learning}, pages
  5960--5969, 2018.

\bibitem{zhou2017critical}
Y.~Zhou and Y.~Liang.
\newblock Critical points of neural networks: Analytical forms and landscape
  properties.
\newblock In {\em International Conference on Learning Representations}, 2018.

\bibitem{zhou2018generalization}
Y.~Zhou, Y.~Liang, and H.~Zhang.
\newblock Generalization error bounds with probabilistic guarantee for {SGD} in
  nonconvex optimization.
\newblock {\em arXiv preprint arXiv:1802.06903}, 2018.

\bibitem{zou2020gradient}
D.~Zou, Y.~Cao, D.~Zhou, and Q.~Gu.
\newblock Gradient descent optimizes over-parameterized deep {ReLU} networks.
\newblock {\em Machine Learning}, 109(3):467--492, 2020.

\end{thebibliography}

\end{document}